\documentclass{article}

\usepackage{microtype}
\usepackage{graphicx}
\usepackage{subfigure}
\usepackage{booktabs} 

\usepackage[accepted]{icml2024-arxiv}

\usepackage{hyperref}

\usepackage{makecell}
\usepackage{tabularx}
\usepackage{multirow}
\usepackage{csquotes}

\usepackage{enumitem}
\setitemize{leftmargin=*, align=left}

\newcolumntype{Y}{>{\centering\fontsize{8}{9}\selectfont\arraybackslash}X}
\newcolumntype{R}{>{\hspace{0.5em}\raggedleft\arraybackslash}X}
\newcolumntype{L}{>{\hspace{0.5em}\raggedright\arraybackslash}X}

\newcommand{\ROLCP}[0]{$ROLC_p$}

\usepackage{amsmath}
\usepackage{amssymb}
\usepackage{mathtools}
\usepackage{amsthm}

\usepackage[capitalize,noabbrev]{cleveref}

\theoremstyle{plain}

\theoremstyle{definition}

\theoremstyle{remark}

\usepackage[textsize=tiny]{todonotes}

\newcommand{\arxivonly}[1]{#1}

\begin{document}

\twocolumn[
\icmltitle{LossVal: Efficient Data Valuation for Neural Networks}

\begin{icmlauthorlist}
\icmlauthor{Tim Wibiral}{ulm,porsche}
\icmlauthor{Mohamed Karim Belaid}{porsche,lmu,idiada}
\icmlauthor{Maximilian Rabus}{porsche}
\icmlauthor{Ansgar Scherp}{ulm}
\end{icmlauthorlist}

\icmlaffiliation{ulm}{Ulm University, Ulm, Germany}
\icmlaffiliation{porsche}{Dr. Ing. h.c. F. Porsche AG Stuttgart, Germany}
\icmlaffiliation{lmu}{Ludwig-Maximilians-Universität Munich, Germany}
\icmlaffiliation{idiada}{IDIADA Fahrzeugtechnik GmbH, Munich, Germany}

\icmlcorrespondingauthor{Tim Wibiral}{tim.wibiral@uni-ulm.de}

\icmlkeywords{Machine Learning, ICML}

\vskip 0.3in
]

\printAffiliationsAndNotice{}  

\begin{abstract}
Assessing the importance of individual training samples is a key challenge in machine learning. Traditional approaches retrain models with and without specific samples, which is computationally expensive and ignores dependencies between data points. 
We introduce LossVal, an efficient data valuation method that computes importance scores during neural network training by embedding a self-weighting mechanism into loss functions like cross-entropy and mean squared error. 
LossVal reduces computational costs, making it suitable for large datasets and practical applications. 
Experiments on classification and regression tasks across multiple datasets show that LossVal effectively identifies noisy samples and is able to distinguish helpful from harmful samples. 
We examine the gradient calculation of LossVal to highlight its advantages. 
The source code is available at: \url{https://github.com/twibiral/LossVal}

\end{abstract}

\section{Introduction} 

Understanding the relative importance of data points is crucial for optimizing model performance, improving model explainability, and guiding the collection of new data~\citep{jia_scalability_2021, molnar_interpretable_2022}. 
This process, known as data valuation, assigns importance scores to every data point.
Applications range from selling or buying data on data markets~\citep{li_theory_2015, baghcheband_shapley-based_2024} to active learning scenarios where acquiring new, high-impact data is costly~\citep{jia_scalability_2021}. 
For example, in passive car safety systems, machine learning models serve as surrogates to predict crash outcomes~\citep{belaid_crashnet_2021, rabus_development_2022}.
Improving these models depends on identifying the most impactful data points, which is challenging due to the presence of both feature and label noise in crash test data. 
Existing data valuation methods struggle to handle both types simultaneously and are computationally expensive.

We propose a novel data valuation approach called \textit{LossVal}, which is both efficient and capable of simultaneously handling feature and label noise. 
Our method takes advantage of the gradient information from standard loss functions by incorporating learnable parameters into the loss function. By dynamically weighting each data point during training, LossVal identifies beneficial and detrimental data points. 
We demonstrate that our method performs comparably to state-of-the-art methods using six classification and six regression datasets from the OpenDataVal benchmark~\citep{jiang_opendataval_2023}. 
In an active learning context, we use LossVal to generate importance scores for a crash test dataset and train a secondary machine learning model to predict these importance scores based on crash configurations. The secondary model is then used to select the next optimal data point for acquisition, allowing us to achieve better model performance while acquiring as few new data points as possible. This effectively reduces the cost of conducting and acquiring new crash data. 
In summary, our contributions are:
\begin{itemize}[topsep=0pt]
    \itemsep1pt
    \item Introduce a self-weighting mechanism for loss functions to compute data importance scores efficiently.
    
    \item Achieve state-of-the-art performance while simultaneously handling label and feature noise.    
  
    \item Use importance scores for data acquisition, especially for costly or hard-to-obtain data like crash tests.
\end{itemize}

\section{Related Work}\label{sec:related_work} 

\arxivonly{
We review relevant literature, focusing on data valuation techniques and machine learning applications in the design of passive car safety systems.

\subsection{Data Valuation}
}

Data valuation, also known as data attribution~\citep{park_trak_2023}, data influence analysis~\citep{hammoudeh_training_2024}, or representer points~\citep{yeh_representer_2018},
aims to assign an importance score to each data point in the training data. 
This score represents how important or influential each data point is for the training of a machine learning model and the model's performance~\citep{ghorbani_data_2019, koh_understanding_2017}. 
There are different approaches to data valuation, and each approach assigns a different meaning or interpretation to the score. Depending on the approach, the importance score is interpreted either as influence~\citep{koh_understanding_2017}, Leave-One-Out (LOO) error~\citep{cook_detection_1977, bae_if_2022}, the Data Shapley value~\citep{ghorbani_data_2019, chen_towards_2019}, or just some form of importance ranking, like the expected utility~\citep{just_lava_2023, kwon_data-oob_2023, yoon_data_2020}. Most approaches assume that the training set is noisy, the test set is clean, and they have access to a clean validation set~\citep{just_lava_2023}.

We divide data valuation into four branches, namely retraining-based approaches (which include LOO, downsampling, and Shapley value methods), gradient-based approaches, data-based approaches (which focus on the data instead of the model), and other approaches that do not fit the first three categories. 
\arxivonly{Most approaches cannot reflect irregular training schedules or shifting data distributions like encountered in reinforcement learning, where the training distribution can shift as the agent's actions improve through training.
In this case, retraining-based or data-based approaches cannot be applied. 
Notable exceptions are Dropout Influence~\citep{kobayashi_efficient_2020}, TracIn~\citep{pruthi_estimating_2020}, and In-Run Data Shapley~\citep{wang_data_2024}.}
We discuss representative methods in each category.
We reflect on further variants, extensions, and applications of data valuation in \Cref{sec:extended-related-work}.
\arxivonly{\Cref{tab:related_work} shows the characterization of some of the most important approaches.

\begin{table*}[t]
\caption{Different approaches to data valuation. Volume-based Data Shapley estimates marginal contribution for different data sources. Shapley Values are optional for  DAVINZ.}\label{tab:related_work}\vskip 0.05in
\begin{center}
\begin{small}
\begin{tabular}{lcccc}
\toprule
Method & \makecell{Shapley\\Values} & \makecell{Needs\\Retraining} & \makecell{Adapts\\Training} & \makecell{Model-\\specific} \\
\midrule
Leave-One-Out~\citep{cook_detection_1977} &  & \checkmark &  &  \\
Influence Functions~\citep{koh_understanding_2017} &  &  & \checkmark & \checkmark \\
Representer Point Selection~\citep{yeh_representer_2018} &  &  & \checkmark & \checkmark \\
Data Shapley~\citep{ghorbani_data_2019} & \checkmark & \checkmark &  &  \\
Influence-Subsample~\citep{feldman_what_2020} &  & \checkmark &  &  \\
D-Shapley~\citep{ghorbani_distributional_2020} & \checkmark &  & \checkmark &  \\
Dropout Influence~\citep{kobayashi_efficient_2020} &  &  & \checkmark & \checkmark \\
KNN-Shapley~\citep{jia_efficient_2020} & \checkmark &  &  & \checkmark \\
TracIn~\citep{pruthi_estimating_2020} &  &  & \checkmark & \checkmark \\
DVRL~\citep{yoon_data_2020} &  &  & \checkmark &  \\
FastIF~\citep{guo_fastif_2021} &  &  & \checkmark & \checkmark \\
Volume-based Data Shapley~\citep{xu_validation_2021} & (\checkmark) &  &  &  \\
Beta Shapley~\citep{kwon_beta_2022} & \checkmark & \checkmark &  &  \\
KNN-Shapley on Embeddings~\citep{jia_scalability_2021} & \checkmark &  &  & \checkmark \\
AME~\citep{lin_measuring_2022} & \checkmark & \checkmark &  &  \\
DAVINZ~\citep{wu_davinz_2022} & (\checkmark) &  & \checkmark & \checkmark \\
DU-Shapley~\citep{garrido-lucero_du-shapley_2023} & \checkmark & \checkmark &  &  \\
LAVA~\citep{just_lava_2023} &  &  &  &  \\
Data-OOB~\citep{kwon_data-oob_2023} &  &  &  & \checkmark \\
TRAK~\citep{park_trak_2023} &  &  & \checkmark & \checkmark \\
Data Banzhaf~\citep{wang_data_2023} &  & \checkmark &  &  \\
In-Run DS~\citep{wang_data_2024} & \checkmark &  & \checkmark & \checkmark \\
Gradient Similarity~\citep{evans_data_2024} &  & \checkmark & \checkmark & \checkmark \\
\midrule
\textbf{LossVal} &  &  & \checkmark & \checkmark \\
\bottomrule
\end{tabular}
\end{small}
\end{center}
\vskip -0.1in
\end{table*}
}

\textbf{Retraining-Based Approaches.}
LOO~\citep{cook_detection_1977} is the simplest form of retraining-based data valuation. 
LOO describes how much the model's performance on the test set would change if a certain training instance had not been part of the training set. 
\arxivonly{It can be calculated by retraining the model $n$ times for $n$ training instances, leaving one of the instances out each time. Some importance scores calculated via LOO may be negative if they lead to a decrease in test performance. With increasing dataset size, the average LOO score shrinks.}
In a similar way, Influence-Subsample~\citep{feldman_what_2020, jiang_opendataval_2023} uses subsampled retraining but estimates the same influence value that Influence Functions compute. 
Data Shapley profits from the benefits of the Shapley value, but is much less efficient~\citep{ghorbani_data_2019, chen_towards_2019}. 
\arxivonly{The Shapley value is a game theoretic concept to calculate the marginal contribution of each player. 
In data valuation, the training instances are the players working together in a coalition, and the payout is the machine learning model's test performance. 
When calculating the so-called Data Shapley value, the model needs to be retrained on all possible coalitions of training instances, i.\,e., all possible subsets. It profits from the mathematical properties of the Shapley value, additivity (or linearity), efficiency, and symmetry~\citep{lloyd_s_shapley_notes_1951, ghorbani_data_2019, molnar_interpreting_2023}. The time complexity for an exact calculation of the Data Shapley values is in $O(2^n)$ for $n$ training instances~\citep{hammoudeh_training_2024}.}
Different methods to efficiently approximate the Data Shapley value have been proposed. Average Marginal Effect (AME)~\citep{lin_measuring_2022} uses linear regression coefficients to approximate the Shapley values, Beta Shapley~\citep{kwon_beta_2022} relaxes the efficiency axiom, DU-Shapley~\citep{garrido-lucero_du-shapley_2023} draws samples from a uniform distribution, and D-SHAPLEY~\citep{ghorbani_distributional_2020} reformulates the Data Shapley to consider the underlying distribution of the data.
\citet{kwon_data-oob_2023} use bagging to train an ensemble of models on different subsets of the same data and estimate importance scores using the Out-of-bag (OOB) error. 
The importance score of a training instance depends on the performance difference between models trained on subsets with and without the instance~\citep{kwon_data-oob_2023}.

\textbf{Gradient-Based Approaches}
utilize training gradients to calculate an importance score. Influence Functions are a staple in statistics for finding influential data points using the Hessian matrix~\citep{cook_detection_1977, cook_characterizations_1980, cook_assessment_1986}. It can be applied to more complex machine learning tasks~\citep{koh_understanding_2017, koh_accuracy_2019} but is computationally expensive and relies on the convexity of the underlying model~\citep{koh_understanding_2017}. Various techniques have been proposed to approximate the exact values or to speed up the calculation of influence functions~\citep{feldman_what_2020, guo_fastif_2021, schioppa_scaling_2022}. 
\arxivonly{The importance score (or influence) estimated by Influence Functions is more similar to the LOO error or the proximal Bregman response function than to the Data Shapley value~\citep{bae_if_2022}.}
Approaches exploiting gradient information include utilizing the generalized representer theorem to find representer points~\citep{yeh_representer_2018}, tracing back gradient updates during training~\citep{pruthi_estimating_2020}, observing gradient changes in lower dimensionality space~\citep{park_trak_2023}, and measuring the similarity between training and validation gradients~\citep{evans_data_2024}. 
Finally, gradient information is also used to estimate Data Shapley values with a single training run~\citep{wang_data_2024}.

The method proposed in this paper also falls into the gradient-based category. 
However, we exploit the gradient information only implicitly during the model training.

\paragraph{Data-Based Approaches.}
The aforementioned approaches rely on machine learning models to estimate an importance score, meaning the importance score is biased towards the model used. Alternative approaches assign a \enquote{model-agnostic}~\citep{xu_validation_2021} importance score to each data point based only on the data. \citet{xu_validation_2021} calculate a volume measure for each data point by considering the diversity of the data, which is correlated with learning performance. \citet{just_lava_2023} optimize a weighted optimal transport distance to calculate the distance between noisy training data and clean validation data, interpreting the distance as a proxy for test performance.
\arxivonly{
By weighting the training instances to minimize the distance between training and validation data, they can maximize the validation performance and assign an accurate valuation to each training instance~\citep{just_lava_2023}.}

\paragraph{Other Approaches.}
Not all approaches fit the previous three categories.
\citet{kobayashi_efficient_2020} identify sub-networks of a neural network that were trained slightly differently resulting from dropout zero-masking. They analyze how different sub-networks perform based on their training data. DAVINZ~\citep{wu_davinz_2022} uses the generalization boundary to estimate how a change in the training data would change the test performance. DVRL applies reinforcement learning to estimate importance scores~\citep{yoon_data_2020}. 
Various approaches use $k$-Nearest-Neighbors (KNN)-based methods to estimate Data Shapley values, as these can be calculated more efficiently with KNN~\citep{jia_efficient_2020, jia_scalability_2021, belaid_optimizing_2023}.

\arxivonly{
\subsection{Machine Learning in Passive Car Safety}
In passive car safety, the focus is on systems that protect vehicle occupants during a crash, such as airbags, belt force limiters, and irreversible belt pretensioners. Unlike active safety systems, which aim to prevent collisions (e.\,g., lane-keeping assistance or emergency braking systems), passive safety features are only triggered once a collision is unavoidable. Ethical concerns surrounding interventions that actively control the vehicle (e.\,g., swerving into oncoming traffic)~\citep{hansson_self-driving_2021} do not apply to passive safety systems, as their purpose is purely protective.

Machine learning techniques have been applied to optimize various parameters of passive safety systems~\citep{belaid_crashnet_2021, liu_damage_2023, mathieu_minimizing_2024, rabus_development_2022, sun_adaptive_2023, rabus_modell_2024}. Key optimizations include the belt force load limiter, airbag vent hole size, and load limiter level switching time. These optimizations are critical for minimizing injury risk during collisions~\citep{rabus_modell_2024}. For example, \citet{belaid_crashnet_2021} employed a convolutional neural network to predict chest acceleration during a crash based on vehicle and restraint system parameters. Similarly, \citet{rabus_development_2022} introduced the Real Occupant Load Criterion (\ROLCP), a metric used to estimate crash severity. Their approach used a combination of machine learning models to predict the \ROLCP{} from vehicle data and restraint system configurations. In another study, \citet{mathieu_minimizing_2024} applied reinforcement learning to find restraint system parameters that resulted in lower occupant loads compared to traditional methods. Furthermore, \citet{sun_adaptive_2023} demonstrated that Gaussian processes can be used to dynamically adjust restraint system parameters based on the occupant's height and weight, further improving occupant protection in real-world accidents.
}

\section{LossVal}\label{sec:lossval-loss}\label{sec:methods} 

The idea of LossVal is to introduce instance-specific weights into the loss function to estimate and update the importance of samples during training.
The proposed loss function LossVal has two factors, an instance-weighted target loss $\mathcal{L}_{w}$ 
and the optimal transport distance $\operatorname{OT}_{w}$, defined as: 

\[ \operatorname{LossVal} = \mathcal{L}_{w}(y, \hat{y}) \cdot \operatorname{OT}_{w}(X_{train}, X_{val})^{2} \,. \]

For the target loss $\mathcal{L}_{w}$, we use instance-weighted formulations of existing loss functions, like a weighted cross-entropy loss or weighted mean-squared error (see Sections~\ref{sec:lossval-in-classification-tasks} and~\ref{sec:lossval-in-regression-tasks}).
The model's prediction is denoted by $\hat{y}$, while $y$ represents the target values. 
The optimal transport distance $\operatorname{OT_{w}}$ takes the features of the training data $X_{train}$ and validation data $X_{val}$ as input.

The weighted formulations of loss functions add learnable weights to the local loss, one weight for each training instance. 
All weights $w_n$ are initialized to $1$.
The model learns to down-weight noisy or less informative data points and up-weight highly informative ones. 
Incorporating the weighted distribution distance $\operatorname{OT_{w}}$ ensures that the feature space is also considered when optimizing the instance-specific weights.

Our intuition for multiplying $\mathcal{L}_{w}$ and $\operatorname{OT_{w}}$ instead of adding them is twofold. First, multiplication means the overall loss becomes zero once the target loss becomes zero.
This stops any further updates of the instance-specific weights. 
Second, when calculating the gradient of $\operatorname{LossVal}$ with respect to $w_j$, one can see that the multiplicative variant leads to a more informative gradient for the instance-specific weights. 
By using multiplication, the weights $w_i$ learned for instance $i$ are also influenced by the other weights $w_j$ with $j\neq i$ during gradient descent. 
This is not the case if addition is used.
We demonstrate this in detail in \Cref{app:lossval_gradients}.

Furthermore, we found that using the squared distance $\operatorname{OT}^2$ leads to better results than using the distance without squaring.
We demonstrate this in the ablation study in \Cref{sec:ablation:components}.

\subsection{Weighted Loss for Classification}\label{sec:lossval-in-classification-tasks}

The cross-entropy loss ($\operatorname{CE}$) is widely used for classification tasks~\citep{wang_comprehensive_2022}.
We incorporate the data valuation into CE by introducing instance-specific weights $w_n$:

\[\operatorname{CE}_{w} = - \sum^{N}_{n=1} \left[ w_{n} \cdot \sum^{K}_{k=1} y_{n,k} \log(\hat y_{n,k}) \right]\,,\]

where $N$ denotes the number of training samples, $K$ denotes the number of classes in the training set, $y_{n,k}$ the true class vector, $\hat y_{n,k}$ is the prediction of the model, and $w_{1}, \dots, w_{N}$ are the instance-specific weights (which are interpreted as the importance scores).

Two key points distinguish $\operatorname{LossVal}$ from other weighted loss functions.
First, the weights are applied per instance as opposed to per class, like in focal loss~\citep{lin_focal_2018}. 
Second, our weights are learnable parameters, optimized during training via gradient descent. This approach bears similarities to self-paced learning~\citep{kumar_self-paced_2010}, which dynamically adjusts the subset of training samples for fitting based on their difficulty.

\subsection{Weighted Loss for Regression}\label{sec:lossval-in-regression-tasks}
For regression, the mean squared error ($\operatorname{MSE}$) is widely used~\citep{wang_comprehensive_2022}.
We incorporate the instance weights into the $\operatorname{MSE}$ similarly to the modification of the cross-entropy loss, 
with $N$ samples, target value $y_n$, predicted value $\hat y_{n}$, 
and instance-specific weights $w_n$.
\[
\operatorname{MSE}_{w} = \sum^{N}_{n=1} w_{n} \cdot (y_{n} - \hat{y}_{n})^2 \,.
\]

Similarly, iteratively reweighted least squares (IRLS)~\citep{holland_robust_1977} is a linear regression technique that dynamically adjusts instance-based weights throughout the optimization process. IRLS primarily aims to down-weight outliers to improve model fit, which differs from the objectives of LossVal. Furthermore, LossVal results in a more complex gradient computation by integrating the optimal transport distance into the $\operatorname{MSE}$.

\subsection{Weighted Optimal Transport}

The target loss $\mathcal{L}_{w}$ (i.\,e., a modified cross-entropy loss or modified MSE) mainly takes into account the label and the models' prediction. 
This means that a weighted target loss mostly adapts the weights based on information present in labels and predictions. 
We involve the distribution of the input features of the data points into the loss through a weighted optimal transport distance $\operatorname{OT}_w$, allowing feature information to guide the optimization of instance-specific weights. 
The optimal transport cost between two distributions ($X_{train}$ and $X_{val}$) is defined as the fastest way to move all points from the source distribution to the target distribution~\citep{cuturi_sinkhorn_2013}.
\[
\operatorname{OT}_w(X_{train}, X_{val}) = \min_{\gamma \in \Pi(w, 1)} \sum_{n=1}^{N}\sum_{j=1}^{J} c(x_n, x_j) \, \gamma_{n,j} \,,
\]
where $\Pi(w, 1)$ is the set of all joint probability distributions $\gamma$ with marginal $w$ for the training set and a uniform marginal for the validation set, ensuring the transport plan respects the instance-specific weights $w$. Each $\gamma$ defines a possible transport plan for moving the training distribution to the validation distribution. 
The optimal transport plan $\gamma*$ is the transport plan that leads to the shortest distance. 
The cost function $c(x_n, x_j)$ denotes the effort of transport, typically the Euclidean distance $\|x_n - x_j\|^2$. $N$ is the number of training data points, and $J$ the number of validation data points.

Sinkhorn's distance adds the entropy $H(\gamma)$ as a regularization term to $\operatorname{OT}$, which makes $\operatorname{OT}$ differentiable and the calculation of $\gamma*$ more computationally efficient~\citep{cuturi_sinkhorn_2013, feydy_interpolating_2019}. 
\citet{just_lava_2023} showed that Sinkhorn's distance can be effectively utilized in the data valuation context, but it would be possible to use any other weighted distributional distance, too. 

By including the weighted OT in the loss function, gradient descent optimizes the weights to decrease the optimal transport distance between the training and validation set. 
Training data points that are more similar (and therefore closer) to the data points in the validation set get up-weighted, while more different data points get down-weighted. 

\section{Experimental Apparatus}\label{sec:evaluation}

\arxivonly{We outline our datasets, procedures, and baselines used. 
We describe the hyperparameter tuning and the measures.}

\subsection{Datasets}

We employ six widely used classification datasets, which are the focus of the OpenDataVal benchmark~\citep{jiang_opendataval_2023}.
The OpenDataVal benchmark does not include predefined regression datasets, so we select six datasets from the CTR-23 benchmark suite~\citep{fischer_openml-ctr23_2023}.
Finally, we employ a crash test dataset consisting of $1,122$ samples from the 
~\citet{NHTSA_Database} and $154$ 
proprietary crash tests provided by a large car manufacturer~\citep{belaid_crashnet_2021, rabus_development_2022} to evaluate LossVal in an active data acquisition setting. 
Details of the datasets are provided in Appendix~\ref{appendix:datasets}.

\subsection{Procedure}
\label{sec:baselines}
We compare LossVal to ten baselines covering different approaches to data valuation. These are
Data Shapley~\citep{ghorbani_data_2019},
Beta Shapley~\cite{kwon_beta_2022},
Leave-One-Out~\citep{cook_detection_1977}, 
KNN-Shapley~\citep{jia_efficient_2020}, 
Data Banzhaf~\citep{wang_data_2023},
AME~\citep{lin_measuring_2022}, 
Infuence Subsample~\citep{feldman_what_2020},
LAVA~\citep{just_lava_2023}, DVRL~\citep{yoon_data_2020}, 
and Data-OOB~\citep{kwon_data-oob_2023}. 
The baselines are selected based on the OpenDataVal benchmark~\citep{jiang_opendataval_2023}.

We run LossVal and the baselines on the tasks from the OpenDataVal benchmark~\citep{jiang_opendataval_2023}, which are Noisy Label Detection, Noisy Feature Detection, Mixed Noise Detection, Point Addition, and Point Removal. 
Additionally, we demonstrate LossVal's effectiveness for active data acquisition using a crash test dataset. 

Many existing data valuation methods rely on repeated model training.
For example, Data-OOB~\citep{kwon_data-oob_2023} trains $1,000$ MLPs, leading to $5,000$ training epochs.
We limit the number of training epochs to ensure a fair comparison of LossVal and the baselines and test LossVal with 5 and 30 training epochs.
LossVal with 5 epochs demonstrates how the method performs when training the model for the same number of epochs as it was trained in the baseline methods. LossVal with 30 epochs demonstrates how LossVal performs when a training run is not restricted to 5 epochs. This is a fairer comparison considering that methods like Data-OOB or LOO train 1000 models (for $5,000$ epochs overall). 
We repeat every experiment 15 times.

\textbf{Noisy Sample Detection Tasks.}
We introduce label noise (where $p\%$ of the labels get mixed), 
feature noise (add Gaussian noise to $p\%$ of samples), or both into $p\%$ of the labels, where $p \in \{5, 10, 15, 20\}$.
We evaluate how well each data valuation method detects noisy points. 
Noisy samples often contain errors or irrelevant information that can mislead the learning algorithm, reducing the overall model performance. 
An effective data valuation method should assign lower importance scores to these noisy samples. 

\textbf{Point Removal and Point Addition Tasks.} 
We test how removing the most valued data points from the training set affects the model performance. 
Removing valued data points should cause model performance to degrade more quickly than random removal.
We start with the complete training set and iteratively remove the $5\%$ top-valued points, starting from $0\%$ to $50\%$ of the points, retraining each time. 
We use $20\%$ noise on the training samples (either noisy labels, noisy features, or mixed noise). 
We use a logistic regression and a linear regression model to evaluate the test performance. 
We use these simpler models as they are less prone to overfitting when the dataset is very small. 

The point addition task starts with $5\%$ training data.
We iteratively add $5\%$ of the least-valued data points, until we reach $50\%$.
The performance of a good data valuation method should increase slower than randomly adding data points.

\textbf{Active Data Acquisition Task.}
The regression crash test dataset is sorted by time and the first $40\%$ is used for training.
The rest of the data points are randomly allocated to $10\%$ validation, $40\%$ acquisition, and $10\%$ test data. 
This emulates the process of acquiring new data, where we only add crash tests from newer car models to the training set.
\arxivonly{Due to the potential danger of injuries from sub-optimally designed restraint systems and the high costs associated with executing new crash tests, there is substantial interest in only adding high-quality data points to the training data and minimizing the number of data points required for improving the performance of the machine learning model.}

First, a crash model is trained to predict the severity of a crash on an occupant. 
Then we employ a secondary model that guides the active data acquisition process by estimating the potential improvement in the crash model's performance when adding new (unseen) data points. 
Details on the procedure and two models involved are described in \Cref{app:procedure.crash}.

\subsection{Hyperparameter Optimization}
We use three different MLP models, one for the classification tasks, one for the regression tasks, and one for the active data acquisition with crash data. 
Using grid search, we optimized the hyperparameters to maximize accuracy and $R^2$-score on the target variable. 
The hyperparameters are described in \Cref{app:hyperparameters}. 
\arxivonly{Afterward, we continued with the other experiments without modifying the hyperparameters.} 
The Adam optimizer~\citep{kingma_adam_2017} is used for all experiments.

For the baseline methods, we used the hyperparameters provided by the OpenDataVal benchmark~\citep{jiang_opendataval_2023}.
For LossVal, we separately tune the learning rates for the classification and regressions tasks.
The best learning rate for both tasks is  $0.01$.

\subsection{Measures}

\textbf{Noisy Sample Detection.} 
\arxivonly{The benchmark tasks Noisy Label, Noisy Feature, and Mixed Noise Detection can be subsumed under noisy sample detection. 
They are based on the assumption that noisy training samples are less important and that an ideal data valuation algorithm would assign a lower data valuation to them.}
We report the noisy sample detection curve and the F1-scores for all methods. 
Both scores are averaged over all datasets and runs. 
The noisy sample detection curves show the proportion of noisy samples detected by subsequently inspecting the data points with a low importance score. 
For better comparability, we also report the average of the curve. 

Further, the balanced F1 score is calculated to see how many of the actual noisy samples the data valuation detected. The F1 score is given for each data valuation method with respect to the noise level. 
Additionally, we report the overall average F1 score per method.

\textbf{Point Addition and Point Removal.} 
We present the test performance curve of removing or adding the most or least valued data points, respectively. Lower curves indicate better data valuation. For better comparability, we also report the average of the curve.

\textbf{Active Data Acquisition.}
To measure the change resulting from adding 1\% additional data points, we retrain the model on the updated dataset. Then we compare the MSE, $R^2$-score, and mean average percentage error (MAPE) of the original model on the test set with the updated model.

\section{Results} 

\arxivonly{We report the results of the benchmark tasks and the active data acquisition experiment.}

\subsection{Noisy Sample Detection}\label{sec:res:noisy_sample_detection}

\begin{figure*}[!th]
\setlength{\abovecaptionskip}{0pt}
  \centering
\includegraphics[width=0.9\textwidth]{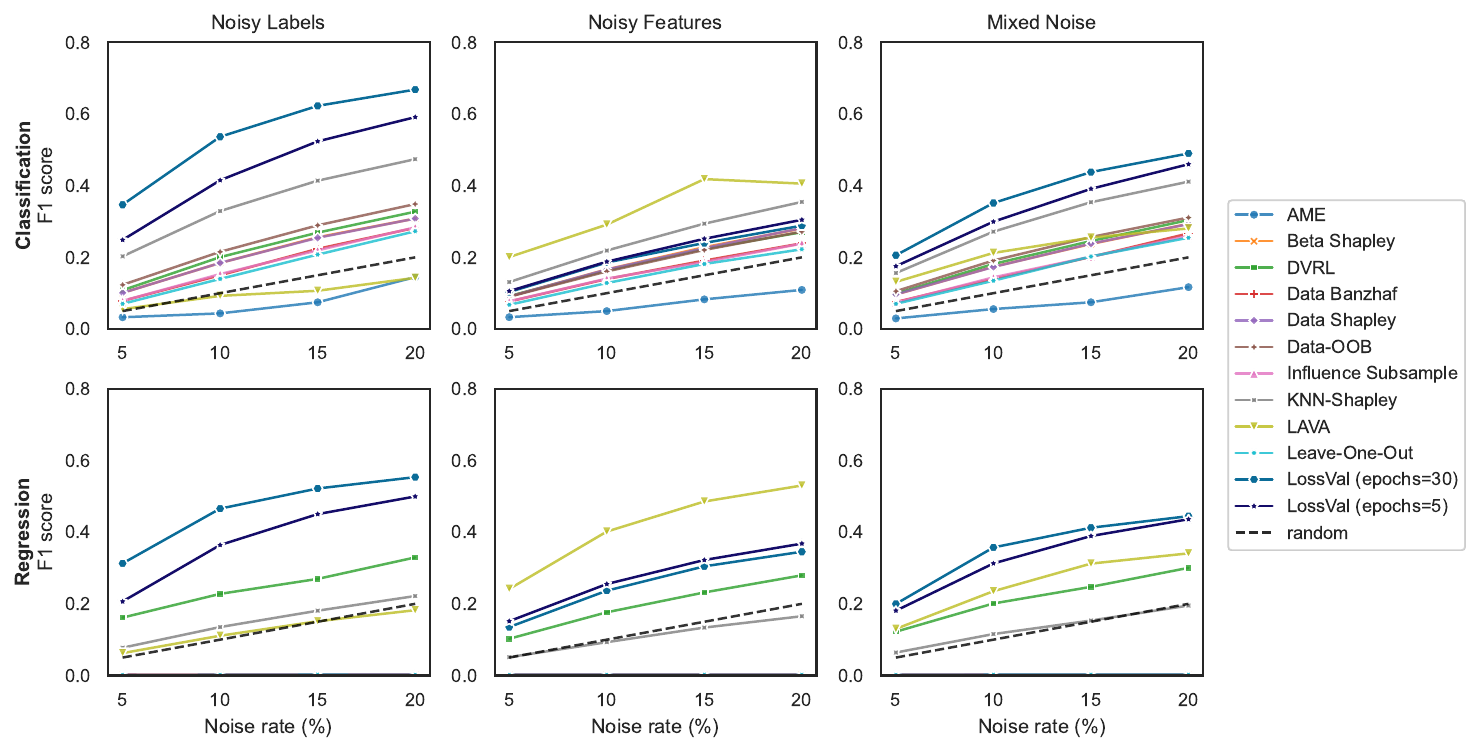}
\vskip -0.15in
  \caption{F1 scores calculated between the set of correct noisy samples and the noisy samples found and averaged over all datasets. Higher is better. 
  In the lower plots, the line graphs of some methods are obscured as they fall almost on the x-axis (very low F1 scores).}\label{fig:exp1:f1_scores}
\end{figure*}

\Cref{fig:exp1:f1_scores} shows how well each data valuation method can find noisy data points. 
The x-axis describes the ratio of noisy samples and the y-axis describes how well the method performs. 
The plots are divided into learning tasks (regression vs. classification) and noise types. 
We observe that 
no single data valuation performs the best for all tasks and noise types. LAVA~\citep{just_lava_2023} outperforms the others in the detection of noisy features. 
Data-OOB~\citep{kwon_data-oob_2023} performs well in detecting noisy labels in classification tasks but struggles with noisy features and regression.
KNN-Shapley~\citep{jia_efficient_2020} shows strong performance in both noisy label and noisy feature detection for classification. DVRL~\citep{yoon_data_2020} shows good performance for both regression and classification but is outperformed by other methods in every task.
LossVal performs well in noisy label and mixed noise detection, even outperforming all other methods for both regression and classification.

\Cref{tab:exp1:f1_scores} shows the average F1 score over all noise levels. LossVal, KNN-Shapley, and LAVA show the best performance in classification tasks. In regression tasks, LossVal outperforms all other methods for mixed noise and noisy label detection but is second after LAVA in noisy feature detection. \Cref{app:sample_detection_curves} gives more profound insight into how well the different approaches can detect noisy samples.

\begin{table*}[!th]
\caption{Average of the noisy sample detection F1 scores of each data valuation method, averaged over all noise rates and datasets. The number after $\pm$ indicates the standard error. Higher is better. 
}\label{tab:exp1:f1_scores}\vskip 0.05in
\begin{minipage}{\linewidth}
\centering\fontsize{8}{9}\selectfont
\begin{tabularx}{0.75\linewidth}{l@{\hskip 0.25em}l|YYY|Y}
\toprule
& &  \makecell{Noisy\\Labels} & \makecell{Noisy\\Features} & \makecell{Mixed\\Noise} &  \makecell{Overall\\Average} \\
\midrule
\multirow{12}{*}{\rotatebox{90}{\textbf{Classification}}}
& AME & 0.074$\pm$.005 & 0.069$\pm$.005 & 0.069$\pm$.005 & 0.071$\pm$.003 \\
& Beta Shapley & 0.212$\pm$.003 & 0.191$\pm$.003 & 0.198$\pm$.003 & 0.201$\pm$.002 \\
& DVRL & 0.226$\pm$.005 & 0.187$\pm$.003 & 0.208$\pm$.004 & 0.207$\pm$.002 \\
& Data Banzhaf & 0.184$\pm$.004 & 0.162$\pm$.004 & 0.171$\pm$.004 & 0.172$\pm$.002 \\
& Data-OOB & 0.244$\pm$.005 & 0.186$\pm$.003 & 0.216$\pm$.003 & 0.215$\pm$.002 \\
& Data Shapley & 0.212$\pm$.003 & 0.191$\pm$.003 & 0.198$\pm$.003 & 0.200$\pm$.002 \\
& Influence Subsample & 0.184$\pm$.004 & 0.161$\pm$.004 & 0.170$\pm$.004 & 0.171$\pm$.002 \\
& KNN-Shapley & \textit{0.355}$\pm$.006 & \underline{0.250}$\pm$.005 & \textit{0.298}$\pm$.005 & \textit{0.301}$\pm$.003 \\
& LAVA & 0.099$\pm$.004 & \textbf{0.329}$\pm$.012 & 0.220$\pm$.008 & 0.216$\pm$.005 \\
& Leave-One-Out & 0.173$\pm$.004 & 0.150$\pm$.004 & 0.166$\pm$.004 & 0.163$\pm$.003 \\
& LossVal (epochs=5) & \underline{0.445}$\pm$.007 & \textit{0.213}$\pm$.003 & \underline{0.332}$\pm$.005 & \underline{0.330}$\pm$.004 \\
& LossVal (epochs=30) & \textbf{0.544}$\pm$.008 & 0.204$\pm$.004 & \textbf{0.371}$\pm$.005 & \textbf{0.373}$\pm$.005 \\

\midrule 

\multirow{12}{*}{\rotatebox{90}{\textbf{Regression}}}
& AME & 0.002$\pm$.000 & 0.002$\pm$.000 & 0.002$\pm$.000 & 0.002$\pm$.000 \\
& Beta Shapley & 0.002$\pm$.000 & 0.002$\pm$.000 & 0.002$\pm$.000 & 0.002$\pm$.000 \\
& DVRL & \textit{0.247}$\pm$.007 & 0.198$\pm$.004 & 0.218$\pm$.005 & 0.221$\pm$.003 \\
& Data Banzhaf & 0.002$\pm$.000 & 0.002$\pm$.000 & 0.002$\pm$.000 & 0.002$\pm$.000 \\
& Data-OOB & 0.002$\pm$.000 & 0.002$\pm$.000 & 0.002$\pm$.000 & 0.002$\pm$.000 \\
& Data Shapley & 0.002$\pm$.000 & 0.002$\pm$.000 & 0.002$\pm$.000 & 0.002$\pm$.000 \\
& Influence Subsample & 0.002$\pm$.000 & 0.002$\pm$.000 & 0.002$\pm$.000 & 0.002$\pm$.000 \\
& KNN-Shapley & 0.154$\pm$.008 & 0.111$\pm$.005 & 0.132$\pm$.006 & 0.132$\pm$.004 \\
& LAVA & 0.127$\pm$.004 & \textbf{0.415}$\pm$.012 & \textit{0.255}$\pm$.008 & \textit{0.265}$\pm$.006 \\
& Leave-One-Out & 0.002$\pm$.000 & 0.002$\pm$.000 & 0.002$\pm$.000 & 0.002$\pm$.000 \\
& LossVal (epochs=5) & \underline{0.380}$\pm$.007 & \underline{0.274}$\pm$.005 & \underline{0.330}$\pm$.006 & \underline{0.328}$\pm$.004 \\
& LossVal (epochs=30) & \textbf{0.464}$\pm$.008 & \textit{0.256}$\pm$.006 & \textbf{0.354}$\pm$.006 & \textbf{0.358}$\pm$.004 \\
\bottomrule
\end{tabularx}
\end{minipage}
\vskip -0.1in
\end{table*}

\subsection{Point Addition and Removal}\label{sec:res:point_add_rem}

\begin{figure*}[!ht]
  \centering
\includegraphics[width=0.9\textwidth]{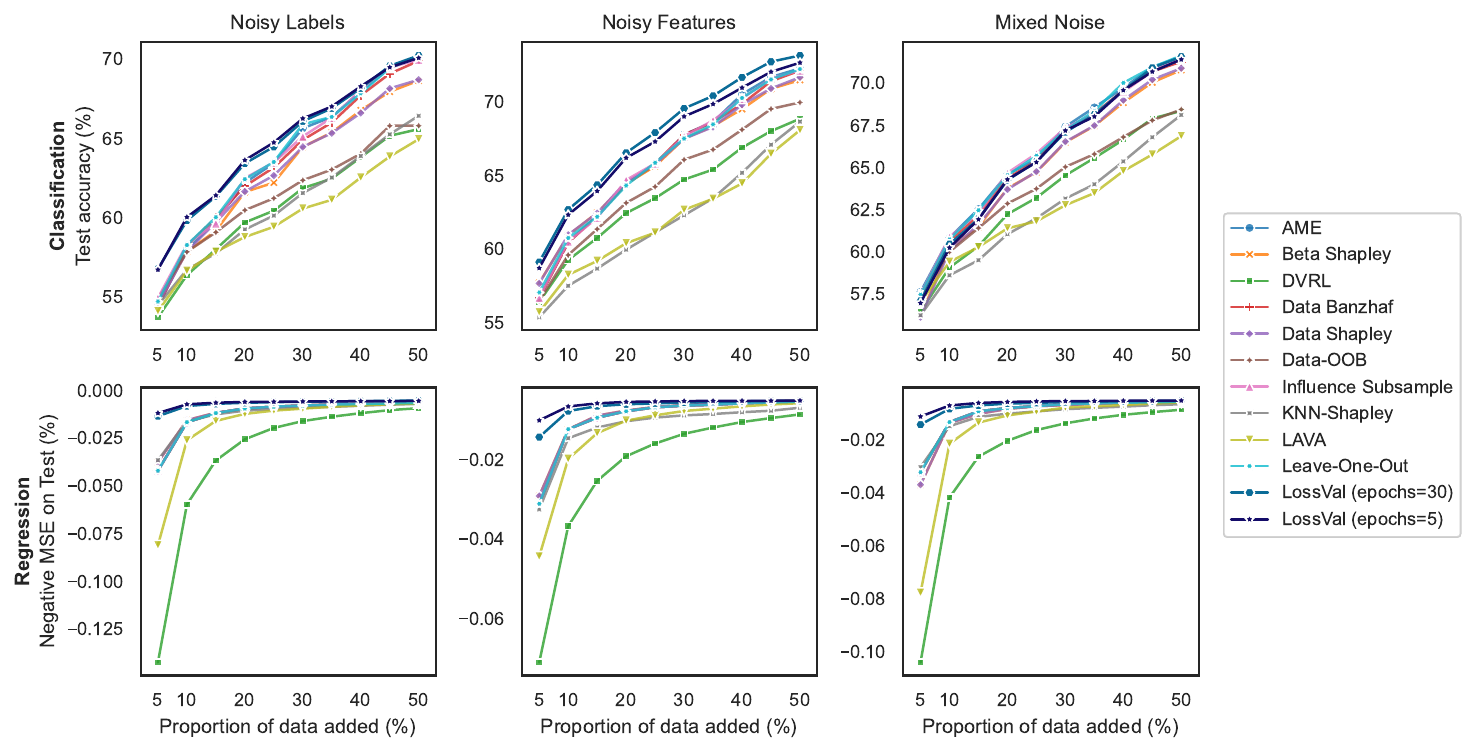}
\vskip -0.15in
  \caption{Adding $x\%$ of data points with low importance score to the training data, averaged over all datasets. A lower curve is better.}\label{fig:exp2:point_addition}
\end{figure*}

\Cref{fig:exp2:point_addition} shows the effect of adding the data points with the lowest importance score to the training set and then retraining the MLP on the updated training set. For regression, we normalized all values per dataset, before averaging over all datasets. Lower curves are better because they indicate a slower increase in test performance when data points with a low importance score are added to the training set. 
For classification, KNN-Shapley performs the best, LAVA comes in second, and DVRL third. For regression, DVRL demonstrated the best performance, followed by LAVA and KNN-Shapley. 
We additionally provide the numerical averages of the curves in \Cref{app:avg_point_removal_addition}.

\begin{figure*}[!ht]
  \centering
\includegraphics[width=0.9\textwidth]{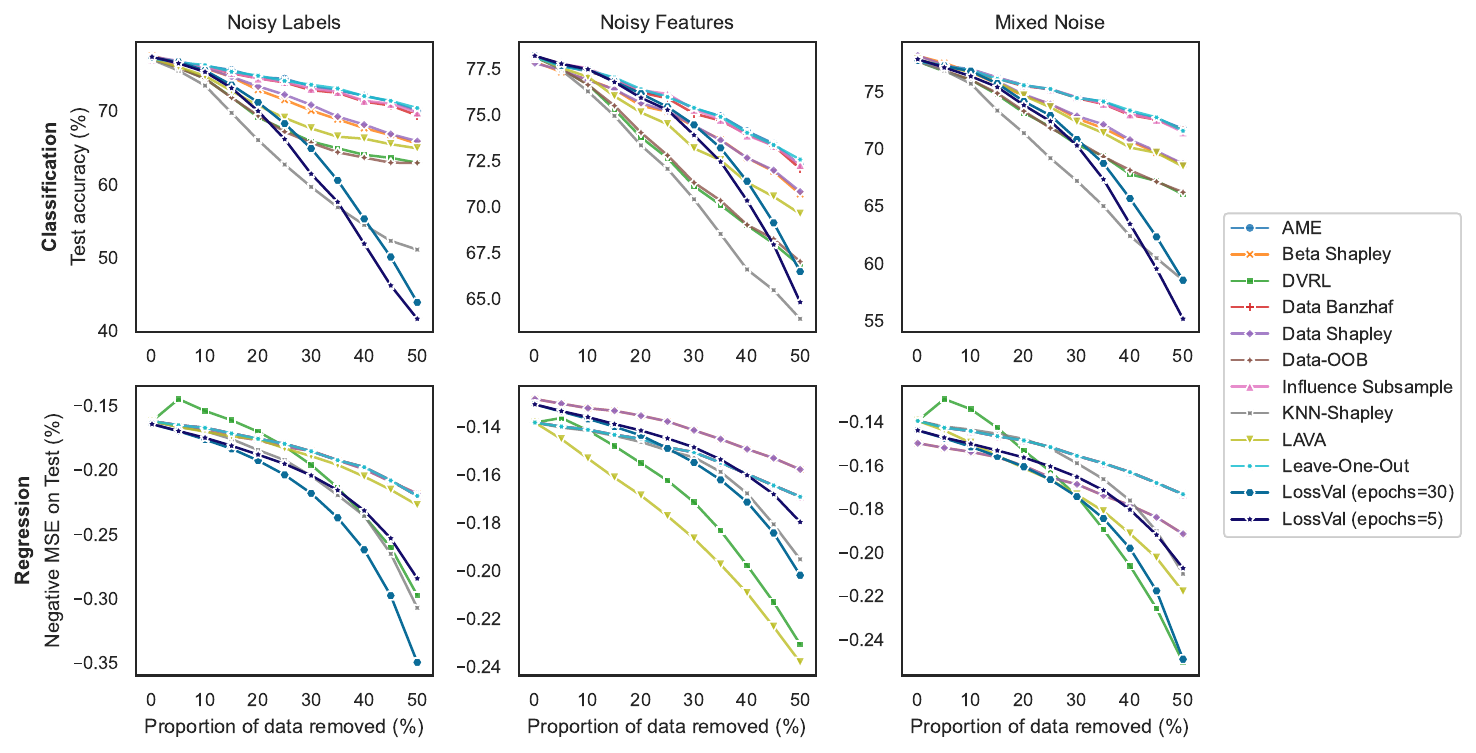}
\vskip -0.15in
  \caption{Removing $x\%$ data points with a high importance score from the training data. A lower curve is better.}\label{fig:exp2:point_removal}
\end{figure*}

The point removal experiment starts from the reverse premise: Removing data points with a high importance score should lead to a steeper decrease in test performance than randomly removing data points. Data valuation methods that lead to a lower point removal curve (compare \Cref{fig:exp2:point_removal}) are better at identifying high-quality data points. In classification, KNN-Shapley achieves the best score, followed by LossVal and DVRL. 
For regression, DVRL performs best, LossVal second best, and LAVA and KNN-Shapley compete for third-best, achieving similar results. The plots show that LossVal performs worse than other methods after removing just a few points, but catches up to the best methods after removing more points.

We found that KNN-Shapley and LossVal outperform all other methods in the point removal experiment. LossVal is better at finding high-quality data points in classification tasks, but KNN-Shapley achieves much better results than all other methods on regression tasks.

\subsection{Active Data Acquisition}\label{sec:res:data_acq}

\begin{table}[ht!]
\caption{Comparison of the test performance in the active data acquisition experiments. The \enquote{baseline} shows the test performance before adding new data, \enquote{random} shows the effect of randomly adding new data.}\label{tab:data_acquisition}\vskip 0.05in
\centering\fontsize{8}{9}\selectfont
\begin{tabularx}{\linewidth}{@{}l|@{}Y@{}Y@{}Y@{}}
\toprule
                      & MSE ($\pm$ SE) & MAPE ($\pm$ SE) & $R^2$ ($\pm$ SE) \\
\midrule
Baseline              &              0.234$\pm$0.008 &     \textbf{0.892}$\pm$0.047 &              0.168$\pm$0.034 \\
Random                &              0.237$\pm$0.007 &              0.935$\pm$0.048 &              0.160$\pm$0.027 \\
\midrule
AME                   &     \textbf{0.231}$\pm$0.009 &  \underline{0.899}$\pm$0.046 &     \textbf{0.182}$\pm$0.033 \\
Beta Shapley          &  \underline{0.232}$\pm$0.008 &              0.925$\pm$0.045 &     \textit{0.175}$\pm$0.034 \\
DVRL                  &              0.242$\pm$0.010 &              0.929$\pm$0.042 &              0.142$\pm$0.042 \\
Data Banzhaf          &              0.246$\pm$0.010 &              0.960$\pm$0.047 &              0.122$\pm$0.044 \\
Data-OOB              &              0.233$\pm$0.010 &              0.930$\pm$0.049 &  \underline{0.175}$\pm$0.036 \\
Data Shapley          &     \textit{0.233}$\pm$0.009 &              0.935$\pm$0.043 &              0.171$\pm$0.036 \\
Influence Subsample   &              0.237$\pm$0.008 &              0.937$\pm$0.046 &              0.159$\pm$0.030 \\
KNN-Shapley           &              0.240$\pm$0.009 &              0.930$\pm$0.049 &              0.149$\pm$0.037 \\
LAVA                  &              0.246$\pm$0.009 &              0.931$\pm$0.043 &              0.122$\pm$0.045 \\
Leave-One-Out         &              0.243$\pm$0.010 &              0.945$\pm$0.045 &              0.137$\pm$0.042 \\
LossVal (epochs = 5)  &              0.242$\pm$0.008 &     \textit{0.912}$\pm$0.042 &              0.137$\pm$0.038 \\
LossVal (epochs = 30) &              0.244$\pm$0.009 &              0.938$\pm$0.048 &              0.131$\pm$0.039 \\
\bottomrule
\end{tabularx}
\vskip -0.1in
\end{table}

In the active data acquisition experiment, we find no strong differences between the different data valuation methods. 
As shown in \Cref{tab:data_acquisition}, AME, Beta Shapley, and Data Shapley improve the $MSE$ the most. 
AME achieves the best $R^2$-score, with Beta Shapley coming second and Data-OOB as third. 

\section{Discussion} 

\arxivonly{We highlight the key findings and implications of our experiments. Further, we discuss the limitations of this work.}

\subsection{Key Results}
Our experiments demonstrate that LossVal matches or outperforms state-of-the-art data valuation methods on the OpenDataVal benchmark tasks. 
LossVal's performance is robust over different types of noise and for both regression and classification tasks and it successfully identifies beneficial and detrimental data points in the active data acquisition task. 
Finally, it demonstrated the best performance for the data acquisition task on the vehicle crash test dataset.

LossVal is highly efficient compared to other data valuation methods, with a time complexity in $O(n + T)$, where $n$ is the dataset size and $T$ represents the complexity of a single training run. In contrast, retraining-based methods like Leave-One-Out (LOO) exhibit a time complexity of $O(n \cdot T)$ 
for Data Shapley), making them impractical for large datasets due to the repeated retraining~\citep{hammoudeh_training_2024}. Gradient-tracking methods such as TracIn~\citep{pruthi_estimating_2020} have a time complexity of $O(n \cdot p)$ where $p$ is the number of model parameters because they require constant gradient tracking across iterations, which adds computational overhead~\citep{hammoudeh_training_2024}. Influence-based approaches like Influence Functions have an $O(n \cdot p)$ complexity by leveraging Hessian approximations~\citep{hammoudeh_training_2024}. Runtimes are reported in \Cref{app:sec:runtime}.

\subsection{Limitations and Threats to Validity}

Despite our method's advantages, several limitations must be considered. First, the importance scores generated by LossVal are less informative than the scores generated by other methods. The LOO and Shapley values quantify whether and by how much a model's test performance improves or decreases if a data point is removed. 
The importance scores of LossVal cannot express this, but an exact quantification is not necessary for most applications.

We have compared LossVal to a range of methods covering all branches of data valuation identified in \Cref{sec:related_work}. 
The comparison to ten baseline methods on 13 datasets provides a comprehensive picture of LossVal's performance.
This robust performance suggests that LossVal can reach comparable results on other datasets as well.

We use a relatively inefficient but well-tested implementation of Sinkhorn's distance, where more efficient implementations make LossVal even faster~\citep{just_lava_2023}. Still, LossVal was faster than all other methods, except LAVA and KNN, which do not train an MLP.

Although Data-OOB is model-agnostic, we observed better performance when using logistic regression as the base model rather than an MLP (see \Cref{app:oob_comparision} for details). However, to ensure a fair comparison across all data valuation methods, we avoided tuning model hyperparameters individually per method.

\section{Ablations}
We perform ablations on the components of LossVal to demonstrate their importance for the performance. 
Further, we investigate how LossVal affects the downstream classification and regression performance.

\subsection{Importance of LossVal Components}\label{sec:ablation:components}

The modified loss formulations in \Cref{tab:exp5:ablation} indicate how the results for LossVal change if parts of the loss function are left out. 
We see that all parts of LossVal are important for the results. 
Furthermore, the multiplication of target loss and distribution distance cannot be replaced by addition. 

\begin{table}[!th]
\caption{Ablation study showing the effect of removing parts of the LossVal loss function on the noisy sample detection. The number after $\pm$ indicates the standard error. Higher is better.}\label{tab:exp5:ablation}\vskip 0.05in
\centering\fontsize{8}{9}\selectfont
\begin{tabularx}{\linewidth}{@{}l@{\hskip 0.25em}l@{\hskip 0.25em}|@{}Y@{}Y@{}Y@{}|@{}Y@{}}
\toprule
& &  \makecell{Noisy\\Labels} & \makecell{Noisy\\Features} & \makecell{Mixed\\Noise} &  \makecell{Overall\\Average} \\
\midrule

\multirow{7}{*}{\rotatebox{90}{\textbf{Classification}}}
& $\operatorname{OT}_{w}$         &              0.146$\pm$.003 &  \underline{0.214}$\pm$.004 &              0.196$\pm$.003 &              0.185$\pm$.002 \\
& $\operatorname{OT}_{w}^2$                              &              0.133$\pm$.003 &              0.160$\pm$.003 &              0.157$\pm$.003 &              0.150$\pm$.002 \\
& $\text{CE}_w$                              &     \textbf{0.546}$\pm$.007 &              0.153$\pm$.004 &  \underline{0.367}$\pm$.005 &  \underline{0.356}$\pm$.005 \\
& $\text{CE}_w + \operatorname{OT}_{w}$          &              0.159$\pm$.003 &     \textbf{0.216}$\pm$.004 &              0.201$\pm$.003 &              0.192$\pm$.002 \\
& $\text{CE}_w + \operatorname{OT}_{w}^{2}$  &              0.115$\pm$.003 &              0.110$\pm$.002 &              0.115$\pm$.003 &              0.113$\pm$.001 \\
& $\text{CE}_w \cdot \operatorname{OT}_{w}$                     &     \textit{0.388}$\pm$.005 &              0.196$\pm$.004 &     \textit{0.298}$\pm$.004 &     \textit{0.294}$\pm$.003 \\
& LossVal                                       &  \underline{0.544}$\pm$.008 &     \textit{0.204}$\pm$.004 &     \textbf{0.371}$\pm$.005 &     \textbf{0.373}$\pm$.005 \\
\midrule 

\multirow{7}{*}{\rotatebox{90}{\textbf{Regression}}}
& $\operatorname{OT}_{w}$                               &              0.117$\pm$.003 &              0.137$\pm$.003 &              0.134$\pm$.003 &              0.129$\pm$.002 \\
& $\operatorname{OT}_{w}^2$                                       &              0.137$\pm$.003 &  \underline{0.253}$\pm$.005 &     \textit{0.217}$\pm$.004 &              0.202$\pm$.002 \\
& $\operatorname{MSE}_w$                                  &     \textit{0.230}$\pm$.005 &              0.124$\pm$.003 &              0.184$\pm$.003 &              0.179$\pm$.002 \\
& $\operatorname{MSE}_w + \operatorname{OT}_{w}$        &              0.142$\pm$.003 &     \textit{0.252}$\pm$.005 &              0.217$\pm$.004 &     \textit{0.203}$\pm$.002 \\
& $\operatorname{MSE}_w + \operatorname{OT}_{w}^{2}$  &              0.099$\pm$.002 &              0.099$\pm$.002 &              0.099$\pm$.002 &              0.099$\pm$.001 \\
& $\operatorname{MSE}_w \cdot \operatorname{OT}_{w}$                   &  \underline{0.395}$\pm$.007 &              0.213$\pm$.005 &  \underline{0.304}$\pm$.005 &  \underline{0.304}$\pm$.004 \\
& LossVal                                        &     \textbf{0.464}$\pm$.008 &     \textbf{0.256}$\pm$.006 &     \textbf{0.354}$\pm$.006 &     \textbf{0.358}$\pm$.004 \\
\bottomrule
\end{tabularx}
\vskip -0.1in
\end{table}

\subsection{Effect of LossVal on Performance}

Using LossVal for training a machine learning model changes the loss function and, therefore, the gradients used to update the model parameters. To better understand how much the modification of the loss changes the model's test performance, we compared the performance of MLPs trained with or without LossVal. We trained an MLP with the same hyperparameters as in previous experiments on both the regression and classification benchmark datasets. We repeated the training 15 times per dataset with a standard target loss (MSE or cross-entropy loss) and the respective LossVal loss, and calculate the accuracy for classification datasets and the $R^2$ score for regression datasets.

We found no strong difference between the test performance of a model trained using a standard loss or using a LossVal loss. We conducted two-tailed $t$-tests between the standard loss functions and LossVal. For classification, we reject the hypothesis that using LossVal reduces the test accuracy compared to using the cross-entropy loss, as no statistically significant difference was found between the two conditions, $t(178)=-0.005,\ p=0.995$. For regression, we also reject the hypothesis that LossVal reduces the test $R^2$ score compared to using the MSE, $t(178)=1.350,\ p=0.179$.

\section{Conclusion}

We introduce LossVal as an effective data valuation method for neural networks. 
It is efficient to train and achieves state-of-the-art results, consistently outperforming state-of-the-art methods in regression. 
Unlike many existing data valuation methods, LossVal maintains robust performance regardless of the noise type or task at hand. 
The promising results and good computational efficiency present LossVal as a viable alternative for data valuation in neural networks. 

LossVal presents several interesting directions for further exploration. It would be valuable to investigate whether it can be successfully extended to different loss functions, such as hinge loss, focal loss, or others. Furthermore, LossVal should be evaluated for applications where efficiency is critical, such as large-scale models and datasets like those used in training large language models. It is challenging to compare results across different methods due to inconsistencies in benchmarks and reporting. Establishing a standard score for data valuation methods would benefit the field and allow meaningful comparisons.

\arxivonly{
\section*{Acknowledgements}

We thank Andor Diera, Jonatan Frank, Marcel Hoffman, Nicolas Lell, Sarah Lingenhöhl, and Helen Lokowandt for their helpful comments and discussions. We acknowledge support by the state of Baden-Württemberg through bwHPC.
}

\section*{Impact Statement}

This paper presents work whose goal is to advance the understanding of data points' importance for training machine learning models. 
There are many potential societal consequences of our work, none of which we feel must be specifically highlighted here.

\bibliography{bibliography}
\bibliographystyle{icml2024}

\newpage

\appendix
\section*{Supplementary Materials}

\section{Gradient Calculation for LossVal}\label{app:lossval_gradients}

We discuss the difference between the gradients resulting from the variant of LossVal where target loss and distribution distance are summed up, i.\,e. $\operatorname{LossVal}^+$, and the variant where they are multiplied, i.\,e. $\operatorname{LossVal}^\bullet$. 
The gradients of the loss with respect to the instance-specific weights are the basis for updating the weights during training. 
Obtaining a better understanding of how the weights are computed improves our intuition about why using the multiplication in $\operatorname{LossVal}^\bullet$ works better.
We also leave out the squaring of the distribution distance $\operatorname{OT}$ for simplicity of the derivatives. 

Computing the gradient for the additive variant
\begin{equation*}
\operatorname{LossVal}^+ = \mathcal{L}_w(y, \hat y) + \operatorname{OT}_w(X_{\text{train}}, X_{\text{test}})
\end{equation*}

with respect to the weight $w_i$ of an instance $i$ is simply
\begin{equation}
\label{eq:additive_lossval}
\frac{\partial  \operatorname{LossVal}^+}{\partial w_i} = \frac{\partial \mathcal{L}_w}{\partial w_i} + \frac{\partial \operatorname{OT}_w}{\partial w_i} \,.
\end{equation}

For the multiplicative variant

\begin{equation*}
\operatorname{LossVal}^\bullet = \mathcal{L}_w(y, \hat y) \cdot \operatorname{OT}_w(X_{\text{train}}, X_{\text{test}})
\end{equation*}

the chain rule is to be applied and results in
\begin{equation}\label{eq:mult_lossval}
\frac{\partial \operatorname{LossVal}^\bullet}{\partial w_i} = \left( \frac{\partial \mathcal{L}_w}{\partial w_i} \cdot \operatorname{OT}_w \right) + \left( \mathcal{L}_w \cdot \frac{\partial \operatorname{OT}_w}{\partial w_i} \right) .
\end{equation}

It is easy to see how the weights $w_j$ for instances $j$ with $j \neq i$ get dropped in the gradient in \Cref{eq:additive_lossval}.
But it persists in the gradient in \Cref{eq:mult_lossval}. 

We briefly have a closer look at $\frac{\partial \mathcal{L}_{w}}{\partial w_{i}}$ and $\frac{\partial \operatorname{OT}_{w}}{\partial w_{i}}$ for the case of using the MSE loss.
$N$ is the number of training samples, and $J$ is the number of validation samples. 
We obtain
\begin{equation*}
\frac{\partial \mathcal{L}_{w}}{\partial w_{i}} = \frac{\partial}{\partial w_{i}} \sum^{N}_{n=1} w_{n} \cdot (y_{n} - \hat{y}_{n})^{2} = (y_{i} - \hat{y}_{i})^{2}
\end{equation*}

and for $\operatorname{OT}$ with the cost function $c(x_{n}, x_{j})$, we obtain
\begin{equation*}
\frac{\partial \operatorname{OT}_{w}}{\partial w_{i}} = \frac{\partial}{\partial w_{i}} \sum_{n=1}^{N}\sum_{j=1}^{J} w_{n} \cdot c(x_n, x_j) = \sum_{j=1}^{J} c(x_i, x_j) \,.
\end{equation*}

Note that we assume that we already found the optimal transportation plan $\gamma*$.

Going back to \Cref{eq:additive_lossval} for $\operatorname{LossVal}^+$, we find that
\begin{align}
\frac{\partial \mathcal{L}_{w}}{\partial w_{i}} + \frac{\partial \operatorname{OT}_{w}}{\partial w_{i}} & = (y_{i} - \hat{y}_{i})^{2} + \sum_{j=1}^{J} c(x_i, x_j) \,. \label{eq:additive_gradient}
\end{align}

For \Cref{eq:mult_lossval}, we find for $\operatorname{LossVal}^\bullet$ that
\begin{align}
 & \left( \frac{\partial \mathcal{L}_w}{\partial w_i} \cdot \operatorname{OT}_w \right) + \left( \mathcal{L}_w \cdot \frac{\partial \operatorname{OT}_w}{\partial w_i} \right) \nonumber \\
= & (y_i - \hat{y}_i)^2 \cdot \operatorname{OT}_w + \mathcal{L}_w \cdot \sum_{j=1}^{J} c(x_i, x_j) \,. \label{eq:mult_gradient}
\end{align}

We observe that for the additive variant $\operatorname{LossVal}^+$ in \Cref{eq:additive_gradient}, the gradient (and therefore the weight updates during training) only depends on the \textit{local} loss for datapoint~$i$ and the \textit{local} optimal transport distance for the datapoint~$i$. 
However, for the multiplicative variant of $\operatorname{LossVal}^\bullet$ in \Cref{eq:mult_gradient}, the gradient depends not only on the local loss and local distance but on the overall loss and the overall optimal transport distance. Here, all instance-specific weights take part in updating each individual weight $w_i$, potentially making the gradient more informative.

\section{Vehicle Crash Tests Background}

Improving the crashworthiness and the restraint systems of a car is fundamental for saving the lives of the occupants and reducing injuries in a collision. To develop optimal passive restraint systems, such as airbags and seatbelts, engineers traditionally rely on physical crash tests or virtual simulations~\citep{rabus_modell_2024}. However, these tests and simulations, while invaluable, are prohibitively expensive.
A single crash test costs hundreds of thousands of dollars, and high-fidelity simulations cost hundreds of dollars each~\citep{spethmann_crash_2009}. The high costs and execution time associated with crash tests and simulations limit the number of them.

To mitigate these challenges, recent advancements have turned to machine learning models as surrogate tools for crash testing~\citep{belaid_crashnet_2021, liu_damage_2023, mathieu_minimizing_2024, rabus_development_2022, sun_adaptive_2023, rabus_modell_2024}. By training models to predict the crash severity based on vehicle parameters, engineers can virtually assess and optimize safety features. Using machine learning models as surrogates for crash tests and simulations allows them to try out more different restraint system configurations and find good solutions faster. However, the effectiveness of these surrogate models is highly dependent on the quality and relevance of the training data used~\citep{budach_effects_2022, chen_data_2021, rabus_modell_2024}.

Publicly available crash test data goes back 40 years~\citep{NHTSA_Database, EuroNCAP_Database}, and progress in car design, materials, and technologies means that older results are not necessarily transferable to modern cars. To improve the machine learning models for current cars and prototypes, we need to identify which crash tests are beneficial and determine if additional training data, like crash tests and simulations, is needed. Understanding the importance of an individual data point for the model's performance is crucial for prioritizing the acquisition of new data that offers the greatest potential for enhancing predictive accuracy.

\section{Details of the Datasets}
\label{appendix:datasets}

\subsection{Classification Datasets}
\Cref{tab:classification_datasets} presents an overview of the six datasets for classification tasks. Those tabular datasets are widely used in literature and are the focus of the OpenDataVal benchmark~\citep{jiang_opendataval_2023}. Each dataset is standardized before use.

\begin{table*}[th!]
\caption{The classification datasets we used. \textit{fried} and \textit{2dplanes} are binarized.}\label{tab:classification_datasets}
\vskip 0.05in
\centering\fontsize{8}{9}\selectfont
\begin{tabularx}{\linewidth}{LYYYYY}
\toprule
Dataset &  \thead{\fontsize{8}{9}\selectfont Sample\\Size} &  \thead{\fontsize{8}{9}\selectfont Input\\Dimension} & \thead{\fontsize{8}{9}\selectfont Number of\\Classes} &  \thead{\fontsize{8}{9}\selectfont Minor Class\\Proportion} & \thead{\fontsize{8}{9}\selectfont Source} \\
\midrule
electricity & 38,474 & 6 & 2 & 0.50 & \citep{gama_learning_2004} \\
fried & 40,768 & 10 & 2 & 0.50 & \citep{friedman_multivariate_1991}\\
2dplanes & 40,768 & 10 & 2 & 0.50 & \citep{breiman_classification_2017}\\
pol & 15,000 & 48 & 2 & 0.37 & \href{https://www.openml.org/d/722}{OpenML-722} \\
MiniBooNE & 72,998 & 50 & 2 & 0.50 & \citep{roe_boosted_2005} \\
nomao & 34,365 & 89 & 2 & 0.29 & \citep{candillier_design_2012} \\
\bottomrule
\end{tabularx}
\vskip -0.1in
\end{table*}

\subsection{Regression Datasets}
The OpenDataVal benchmark does not include predefined regression datasets, so we selected six datasets from the CTR-23 benchmark suite~\citep{fischer_openml-ctr23_2023} according to specific criteria. We ensured that all selected datasets contain only numeric features, have no missing values, and include at least $4,100$ samples ($1,000$ for training, $100$ for validation, and $3,000$ for testing). Additionally, for datasets with fewer than 45 features, we limited the maximum number of samples to $10,000$. The resulting six regression datasets are similar in numbers of features and samples to the classification datasets, as described in \Cref{tab:regression_datasets}. Like the classification datasets, these were standardized before use.

\begin{table*}[th!]
\caption{Description of a subset of the regression datasets from the CTR23 benchmark suite we used~\cite{fischer_openml-ctr23_2023}.}\label{tab:regression_datasets}\vskip 0.05in
\centering\fontsize{8}{9}\selectfont
\begin{tabularx}{\linewidth}{LYYYYY}
\toprule
Dataset &  \thead{\fontsize{8}{9}\selectfont Sample\\Size} &  \thead{\fontsize{8}{9}\selectfont Input\\Dimension} & \thead{\fontsize{8}{9}\selectfont Mean} &  \thead{\fontsize{8}{9}\selectfont Standard\\Deviation} & \thead{\fontsize{8}{9}\selectfont OpenML\\ID} \\
\midrule
           kin8nm &  8,192 & 8 &  0.71 &   0.26 & \href{https://www.openml.org/d/44980}{44980} \\
      white\_wine &  4,898 & 11 &  5.88 &  0.89 & \href{https://www.openml.org/d/44971}{44971} \\
    cpu\_activity &  8,192 & 21 & 83.97 &  18.40 & \href{https://www.openml.org/d/44978}{44978} \\
      pumadyn32nh &  8,192 & 32 & 0 &  0.04 & \href{https://www.openml.org/d/44981}{44981} \\
     wave\_energy &  72,000 &   48 & 3,760,135 & 112,145 & \href{https://www.openml.org/d/44975}{44975} \\
superconductivity & 21,263 & 81 & 34.42 &  34.25 & \href{https://www.openml.org/d/44964}{44964} \\
\bottomrule
\end{tabularx}
\vskip -0.1in
\end{table*}

\subsection{Crash Test Dataset}\label{sec:crashtestdata}
We use a dataset with vehicle crash tests to evaluate the effectiveness of LossVal in active data acquisition. This dataset, created to support the development of restraint systems for vehicles, consists of $1,122$ publicly available crash tests from the National Highway Traffic Safety Administration (NHTSA)~\citep{NHTSA_Database} and $154$ proprietary crash tests provided by Porsche AG~\citep{belaid_crashnet_2021, rabus_development_2022}. For this study, we focus on full-frontal crash tests conducted at $56$ km/h (about $15.6$ m/s). The data contains numerous metrics and sensor data, from which we extract the features described in \Cref{appendix:crash:features}. All features derive from vehicle information or sensors built into the car. 

Our goal is to predict the injury severity for the occupant (in our case, the dummy) from car-bound information alone (without any information from the dummy). The target variable is the Real Occupant Load Criterion (\ROLCP), an adapted variant of the Occupant Load Criterion (OLC)~\citep{rabus_development_2022, rabus_modell_2024}. The \ROLCP{} is calculated from the dummy chest acceleration and highly correlated with the load on the dummy during the crash test. Our goal is to predict the \ROLCP{} using only car-specific features without knowing the acceleration signals of the dummy.

According to the \ROLCP-Model, a vehicle crash can be divided into three phases, as shown in \Cref{fig:crash_example} in \Cref{app:crash_sample}. After impact, the car decelerates (between $0$ and $t_1$ on the time axis), but the dummy does not decelerate immediately because the dummy and car are not rigidly connected. There is some space between the belt and the dummy chest. As the car decelerates, the dummy continues moving at the original speed until the dummy is connected to the vehicle deceleration via the restraint system. The moment of coupling is called $t_1$. The dummy speed at $t_1$ is equal to $v_1$. Between $t_1$ and $t_2$, the dummy experiences a deceleration. $t_2$ is defined as the moment when driver and car speed are equal with a constant rebound speed ($v_2$). The \ROLCP{} is defined as the absolute slope of a line from point A($t_1, v_1)$ to point B($t_2, v_2$), measured in $g$. We refer to \citep{rabus_development_2022, rabus_modell_2024} for a more extensive discussion of the \ROLCP{}.

\subsection{Example, Features, and Configurable Parameters of a Crash Test}\label{app:crash_sample}

\begin{figure}
\setlength{\abovecaptionskip}{0pt} 
  \centering
  \includegraphics[width=\linewidth]{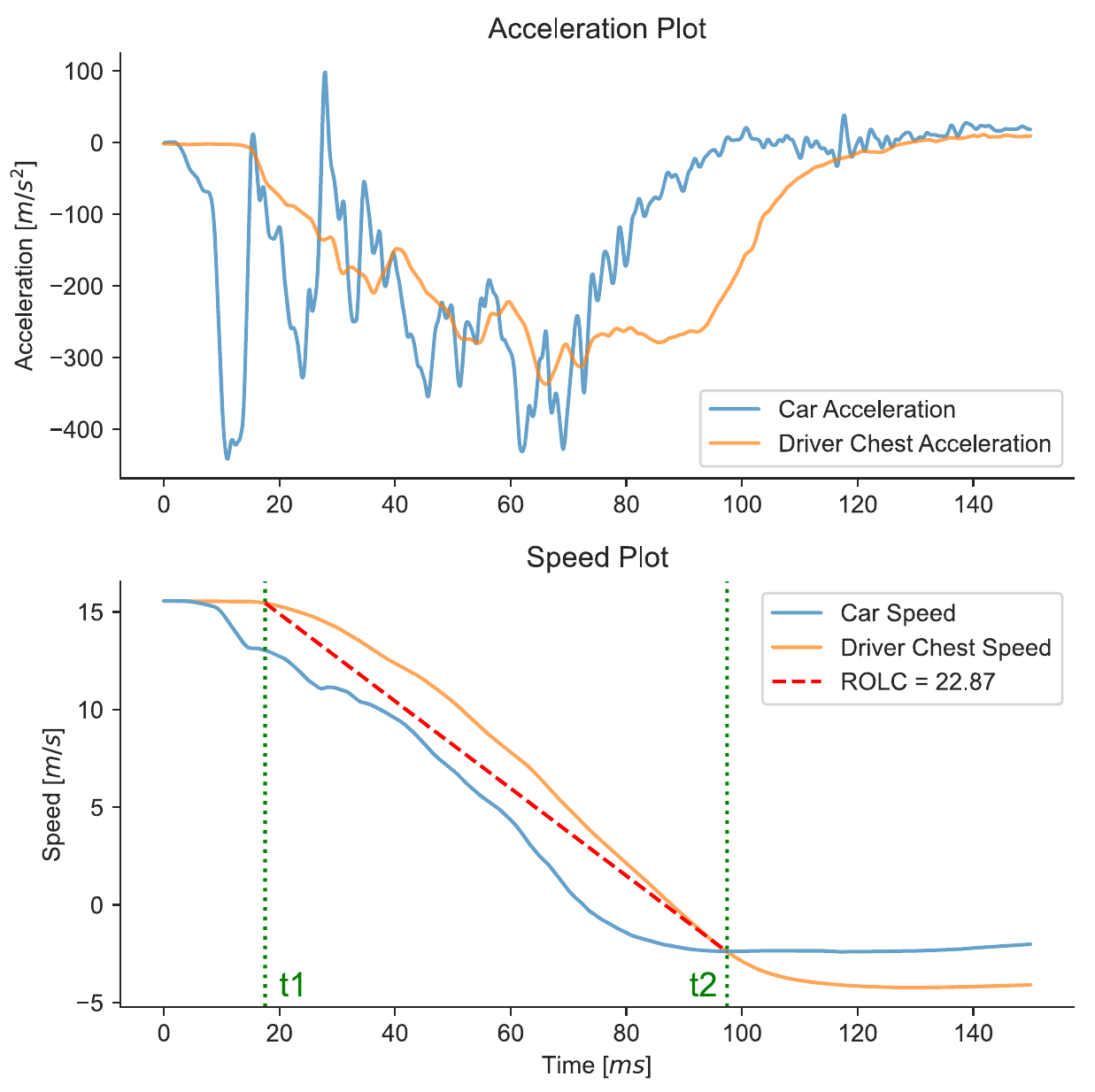}
  \caption{Exemplary crash test showing vehicle and occupant acceleration (top) and speed (bottom), as well as the corresponding \ROLCP model.}
  \label{fig:crash_example}
\end{figure}

\paragraph{Example}
\Cref{fig:crash_example} shows the sensor signals from an exemplary crash test.

\paragraph{Features}
The features of a crash test are in detail:
\label{appendix:crash:features}

\begin{description}[topsep=0pt]
    \itemsep1pt
    \item[Car acceleration.] The car acceleration signal over $130$ ms, sampled every $2$ ms.
    \item[Car body type.] One-hot encoding of the car body type (selection from: convertible, pickup truck, four-door sedan, van, five-door hatchback, utility vehicle, three-door hatchback, two-door coupe, two-door sedan, extended cap pickup, minivan, four-door pickup, station wagon, truck, three-door coupe). 
    \item[Car mass.] Mass of the vehicle in kg.
    \item[Car speed at $t0$.] The speed of the car at the moment of impact. It slightly varies, but is always around $15.6$ m/s.
    \item[Restraint system time-to-fire.] How many milliseconds after impact, the restraint system components (airbag, belt tensioner) fire.
    \item[Chest to steering wheel distance.] The distance between steering wheel and driver chest.
    \item[Number of shoulder belt force limiters.] Either $0$, $1$, or $2$.
    \item[Shoulder belt force level 1.] Threshold of the first level belt force limiter.
    \item[Shoulder belt force level 2.] Threshold of the second level belt force limiter.
    \item[Shoulder belt force limiter switching time.] The point in time when switching from the first to the second belt force limiter.
    \item[Availability of the shoulder belt pretensioner.] Either $1$ when a pretensioner is available or $0$, otherwise.
    \item[Average car acceleration.] Average car acceleration considering only the x-axis (the driving direction).
    \item[Maximum car acceleration.] Maximum car acceleration considering only the x-axis.
    \item[Maximum car acceleration over 3ms.] Maximum car acceleration considering only the x-axis and only accelerations endured for longer than 3 ms (flattening high peaks).
    \item[Dynamic deformation.] Maximum dynamic deformation~\citep{huang_study_1995}.
    \item[Kinetic energy.] The kinetic energy of the car on impact.
    \item[SM\textsubscript{25ms}.] Sliding mean over 25 ms~\citep{gu_structural_2005}.
    \item[TTZV.] Time to zero velocity~\citep{viano_assessing_1990}.
    \item[OLC.] Occupant Load Criterion~\citep{kubler_frontal_2009}.
    \item[OLC++.] Linear combination of OLC, SM\textsubscript{25ms} and TTZV~\citep{kubler_frontal_2009}.
    \item[MCD.] Mean crash deceleration~\cite{rabus_modell_2024}.
    \item[$\boldsymbol\Delta \boldsymbol{V}$.] Maximum velocity difference~\citep{wu_optimization_2002}.
    \item[Rebound velocity.] Maximum rebound speed after impact.
\end{description}

\paragraph{Configurable Parameters}\label{appendix:crash:configurable_features}

In the following, we give the features used for training the secondary model in the active data acquisition experiments. We limit the features used by the secondary model because we want to simulate a guided data acquisition process, where new crash tests are executed. Of course, before a new crash test is executed, we do not know the occupant load. This includes, for example, the weight of the car and the belt force limiter, so we can't use them for predicting the expected value of the crash test. We use only the following features when training the secondary model:

\begin{itemize}[topsep=0pt]
    \itemsep1pt
    \item Car body type.
    \item Car mass.
    \item Restraint system time-to-fire.
    \item Chest to steering wheel distance.
    \item Number of shoulder belt force limiters.
    \item Shoulder belt force level 1.
    \item Shoulder belt force level 2.
    \item Availability of the shoulder belt pretensioner.
\end{itemize}

\section{Detailed Procedure of the Active Data Acquisition Task}
\label{app:procedure.crash}

\begin{figure*}[!tbh]
  \centering
  \includegraphics[width=\textwidth]{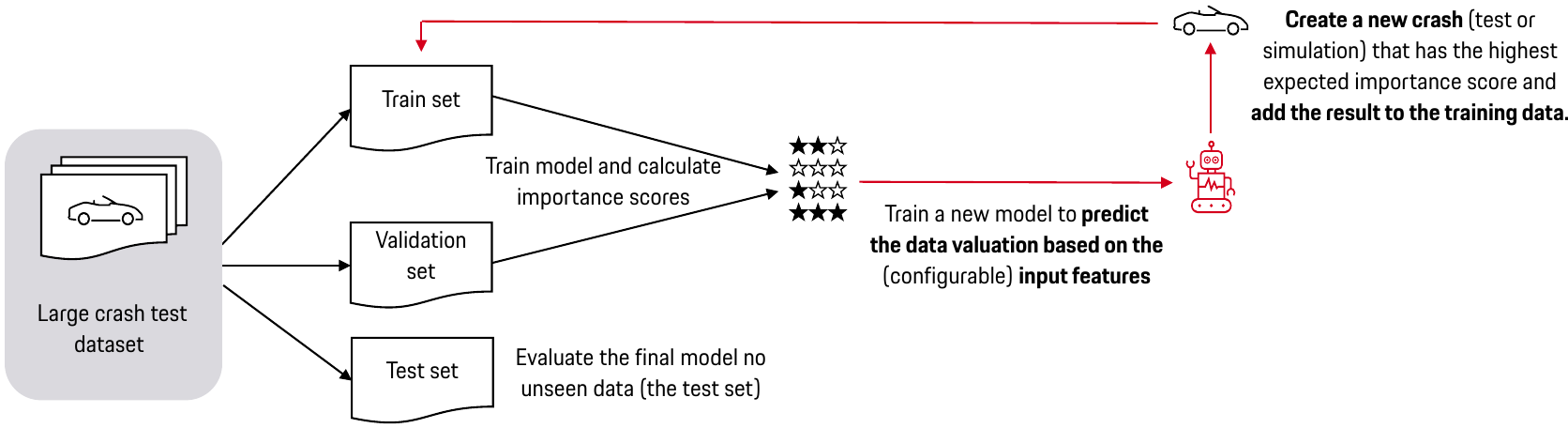}
  \vskip -0.2in
  \caption{Experimental setup for the active data acquisition.}\label{fig:exp4:active_data_acq}
\end{figure*}

We provide details about the active data acquisition experiment using the crash test dataset.
The process is shown in \Cref{fig:exp4:active_data_acq}. 
Since it is not feasible to generate new crash tests for this study, we simulate the data acquisition process using the existing dataset, by taking the highest-expected value data point from an unseen acquisition set.

First, we train a crash model (the MLP optimized for the crash test dataset) on the training data to predict the \ROLCP. Then, we use LossVal and the baseline data valuation methods to estimate importance scores for all training samples. For each method, we train a secondary model (random forest) to predict the importance score based on the features of the corresponding training sample. The secondary model cannot \enquote{see} all the features of the data, but only the \textit{configurable} features. This means the features that are known or can be changed before a crash test is executed, including, for example, the weight of the car and the belt force limiter (the full list is given in \Cref{appendix:crash:configurable_features}). This procedure allows us to simulate an active learning approach that prioritizes data points based on their contribution to improving the model's performance.

The secondary model is used to predict the expected importance scores of the data points in the acquisition set. We take the 1\% data points with the highest expected importance scores from the acquisition set and add them to the training set. Then we train the crash model again with the extended training set and compare the test performance of the model before and after adding more training data. The training of the crash model is repeated 10 times each to reduce the effects of randomness, but the random forest is only fitted once. The whole procedure is repeated 15 times.

\section{Hyperparameters}
\label{app:hyperparameters}

\subsection{Hyperparameter Search Space for the MLP}\label{appendix:hyperparameters:mlp}
\begin{description}[topsep=0pt]
    \itemsep1pt
    \item[Number of hidden layers:] $1, 2, 3, 4, 5$.
    \item[Size of hidden layers:] $10, 20, \dots, 100$.
    \item[Learning rate:] $0.001, 0.01, 0.1$.
    \item[Batch size:] $32, 64, 128$.
    \item[Activation function:] tanh, sigmoid, ReLU~\citep{arras_explaining_2017}
\end{description}

\subsection{Overview of other Hyperparameters}
We collect all hyperparameters used in the experiments here for reference.

\paragraph{\textbf{OpenDataVal Benchmark Experiments:}}
\begin{description}[topsep=0pt]
    \itemsep1pt
    \item[Training / Validation / Test split:] 1000 / 100 / 3000 samples.
    \item[Noise rates:] 5\%, 10\%, 15\%, 20\%.
    \item[Gaussian noise parameters:] $\mu = 0, \sigma=1.0$.
    \item[Number of models for AME:] 1000.
    \item[Number of models for Data-OOB:] 1000.
    \item[Number of training epochs for DVRL:] 2000.
    \item[Number of neighbors $k$ for KNN-Shapley:] 100.
    \item[Number of experiment repetitions:] 25.
    \item[Number of training epochs:] 5 (exception: LossVal is trained for 5 and for 30 epochs).
    \item[MLP hyperparameters:] See \Cref{tab:hyperparameters}.
\end{description}
\vspace{5pt}

\paragraph{\textbf{Active Data Acquisition Experiment:}}
\begin{description}[topsep=0pt]
    \itemsep1pt
    \item[Dataset size:] 1276 samples.
    \item[Training / Validation / Acquisition / Test split:] 40\% / 10\% / 40\% / 10\% of the dataset.
    \item[Newly acquired samples added to training set:] 1\% of the acquisition set. 
    \item[Number of experiment repetitions:] 50.
    \item[MLP training repetitions:] 10.
    \item[Secondary model:] Random forest regressor with 100 estimators.
    \item[Number of models for AME:] Equal to the training set size.
    \item[Number of models for Data-OOB:] Equal to the training set size.
    \item[Number of training epochs for DVRL:] Equal to 2 times the training set size.
    \item[Number of neighbors $k$ for KNN-Shapley:] 10\% of the training set size. 
    \item[Number of training epochs:] 5 (exception: LossVal is trained for 5 and for 30 epochs).
    \item[MLP hyperparameters:] See \Cref{tab:hyperparameters}.
\end{description}

\subsection{Optimal MLP Hyperparameter Values}

The optimal MLP hyperparameter configurations are given in \Cref{tab:hyperparameters} and used in all experiments.

\begin{table}[ht!]
\caption{The MLP hyperparameters we found work best for the three different tasks.}\label{tab:hyperparameters}\vskip 0.05in
\centering\fontsize{8}{9}\selectfont
\begin{tabularx}{\linewidth}{@{}l@{\hskip 0.25em}Y@{}Y@{}Y@{}}
\toprule
&  \thead{\fontsize{8}{9}\selectfont Classification\\ \fontsize{8}{9}\selectfont Benchmark} &  \thead{\fontsize{8}{9}\selectfont Regression\\ \fontsize{8}{9}\selectfont Benchmark} & \thead{\fontsize{8}{9}\selectfont Crash\\ \fontsize{8}{9}\selectfont Scenario} \\
\midrule
Size of hidden layers & 100 & 90 & 100 \\
Number of hidden layers & 5 & 3 & 3 \\
Activation function & ReLU & tanh & tanh \\
Learning rate & 0.1 & 0.01 & 0.01 \\
Batch size & 128 & 32 & 32 \\
\bottomrule
\end{tabularx}
\vskip -0.1in
\end{table}

\section{Extended Results}\label{app:extended_results}

\subsection{Noisy Sample Detection Curves}\label{app:sample_detection_curves}

\begin{figure*}
\setlength{\abovecaptionskip}{0pt}
  \centering
\includegraphics[width=\textwidth]{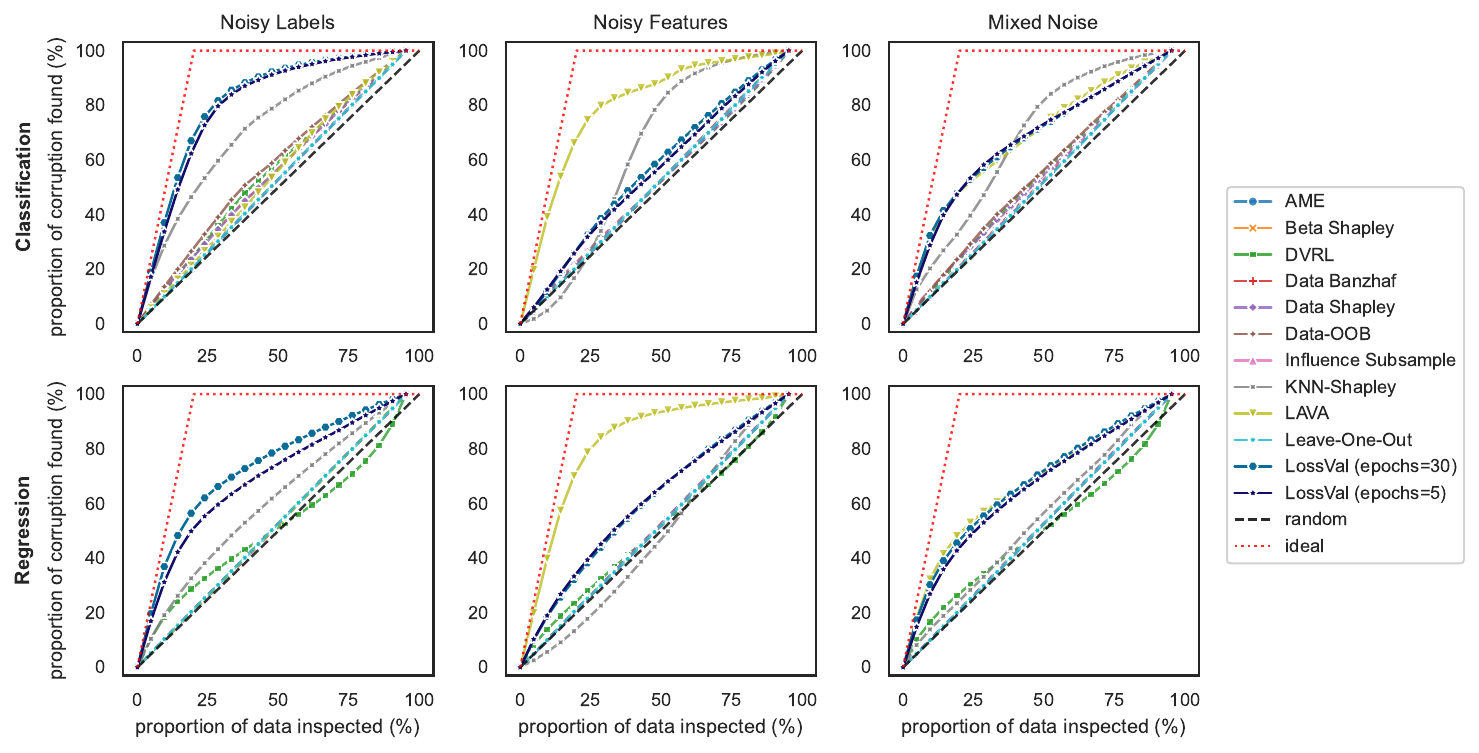}
  \caption{Noisy sample detection for classification (top row) and regression (bottom row). The curves show the average for all classification and regression datasets, respectively. Higher is better. In the lower plot, certain methods are obscured along the random line.}\label{fig:app:noisy_sample}
\end{figure*}

\Cref{fig:app:noisy_sample} shows the noisy sample detection curves from the noisy sample detection experiment. The curve shows the proportion of noisy samples detected per proportion of data inspected, with the underlying assumption that noisy data points will receive the lowest importance scores. Say, we add noise to 20\% of the data points. Then a perfect data valuation method would therefore have detected 25\% of the noisy data points, after inspecting 5\% of all data points (starting with the data points with the lowest importance score). \Cref{tab:app:noisy_sample} gives the average over each curve, dataset, and noise rate. 

\Cref{tab:app:f1_scores_classification} and \Cref{tab:app:f1_scores_regression} show the noisy label detection F1 scores from \Cref{sec:res:noisy_sample_detection} broken down by dataset. They largely reflect the results from \Cref{tab:exp1:f1_scores}.

\subsection{Average of the Point Addition and Removal Curves}\label{app:avg_point_removal_addition}
For better comparability, we give the averages of all the curves in \Cref{tab:exp2:point_addition}. For regression, the negative MSE was normalized by dividing all values by the maximum value. This makes the table more readable because the experiment resulted in very large negative MSE values. Lower values are better because they indicate a faster decrease in test performance when data points with a high importance score are removed from the training set.

\subsection{Importance Score Distribution}
\begin{figure}
\setlength{\abovecaptionskip}{0pt}
  \centering
\includegraphics[width=\linewidth]{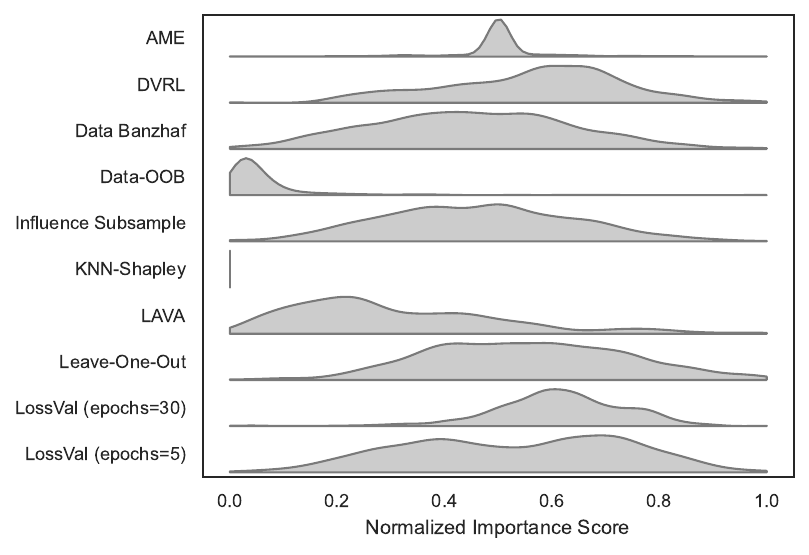}
  \caption{This plot shows the density of the normalized importance scores of each method.}\label{fig:app:importance_scores_density}
\end{figure}

\begin{figure}
\setlength{\abovecaptionskip}{0pt}
  \centering
\includegraphics[width=\linewidth]{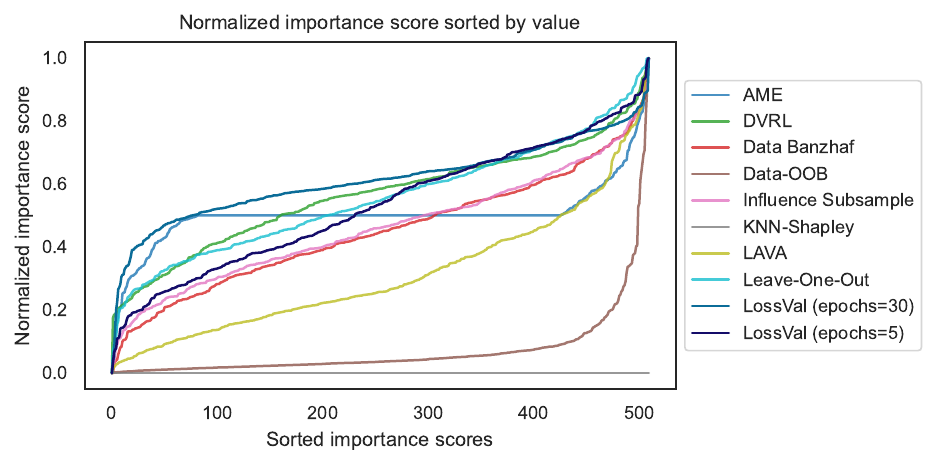}
  \caption{This plot shows the normalized importance scores sorted for each method. The y-axis is the value of the importance score. They are sorted along the x-axis.}\label{fig:app:importance_scores_sorted}
\end{figure}

\Cref{fig:app:importance_scores_density} shows how the importance scores are distributed for each method. KNN-Shapley did fail to find useful importance scores. \Cref{fig:app:importance_scores_sorted} shows the normalized value of the importance scores, when sorted by the value. 

\subsection{Data-OOB Comparison}\label{app:oob_comparision}
\begin{figure*}
\setlength{\abovecaptionskip}{0pt}
  \centering
\includegraphics[width=\textwidth]{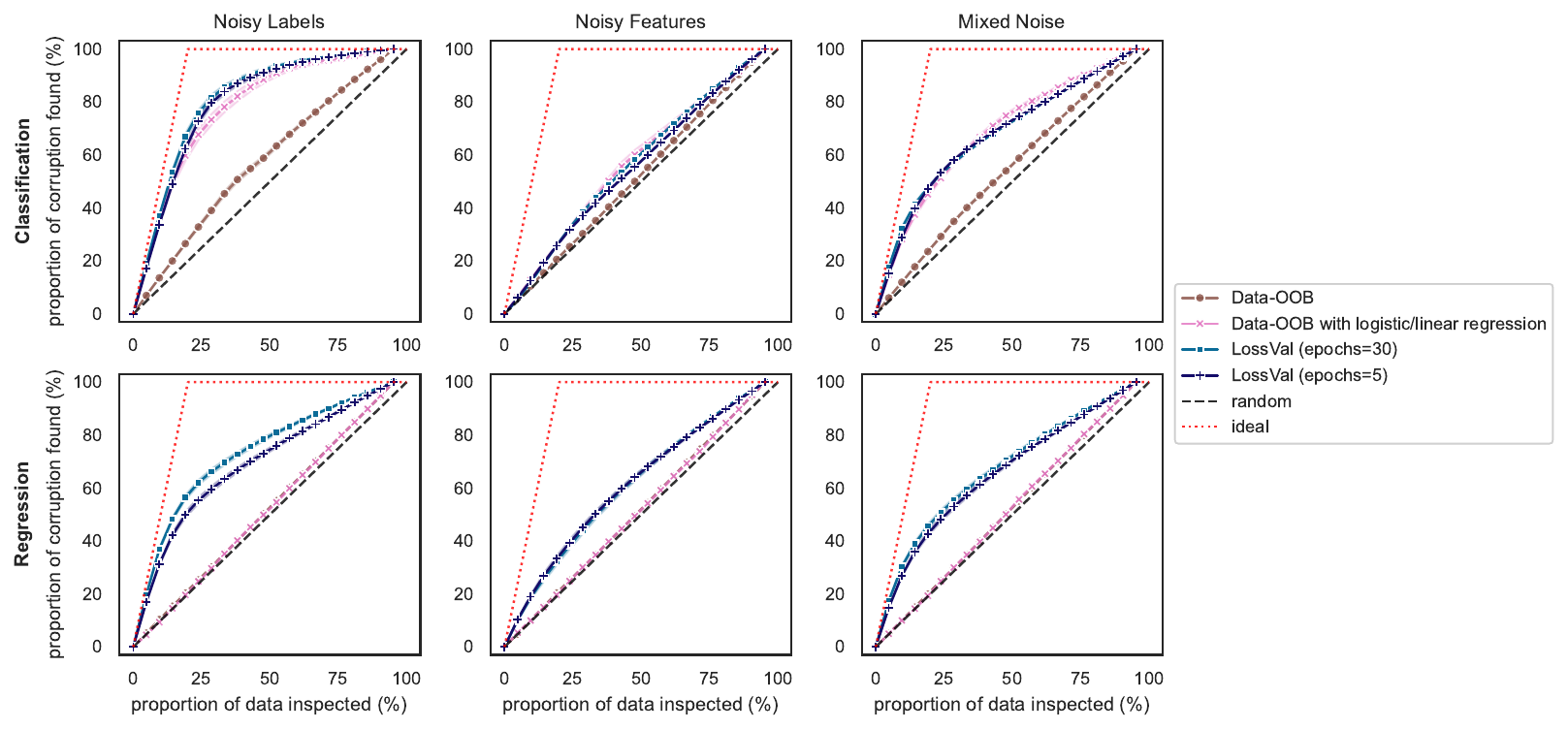}
  \caption{Noisy sample detection for classification (top row) and regression (bottom row). The curves show the average for all classification and regression datasets, respectively. Higher is better.}\label{fig:app:noisy_sample_oob}
\end{figure*}

\begin{figure*}
\setlength{\abovecaptionskip}{0pt}
  \centering
\includegraphics[width=\textwidth]{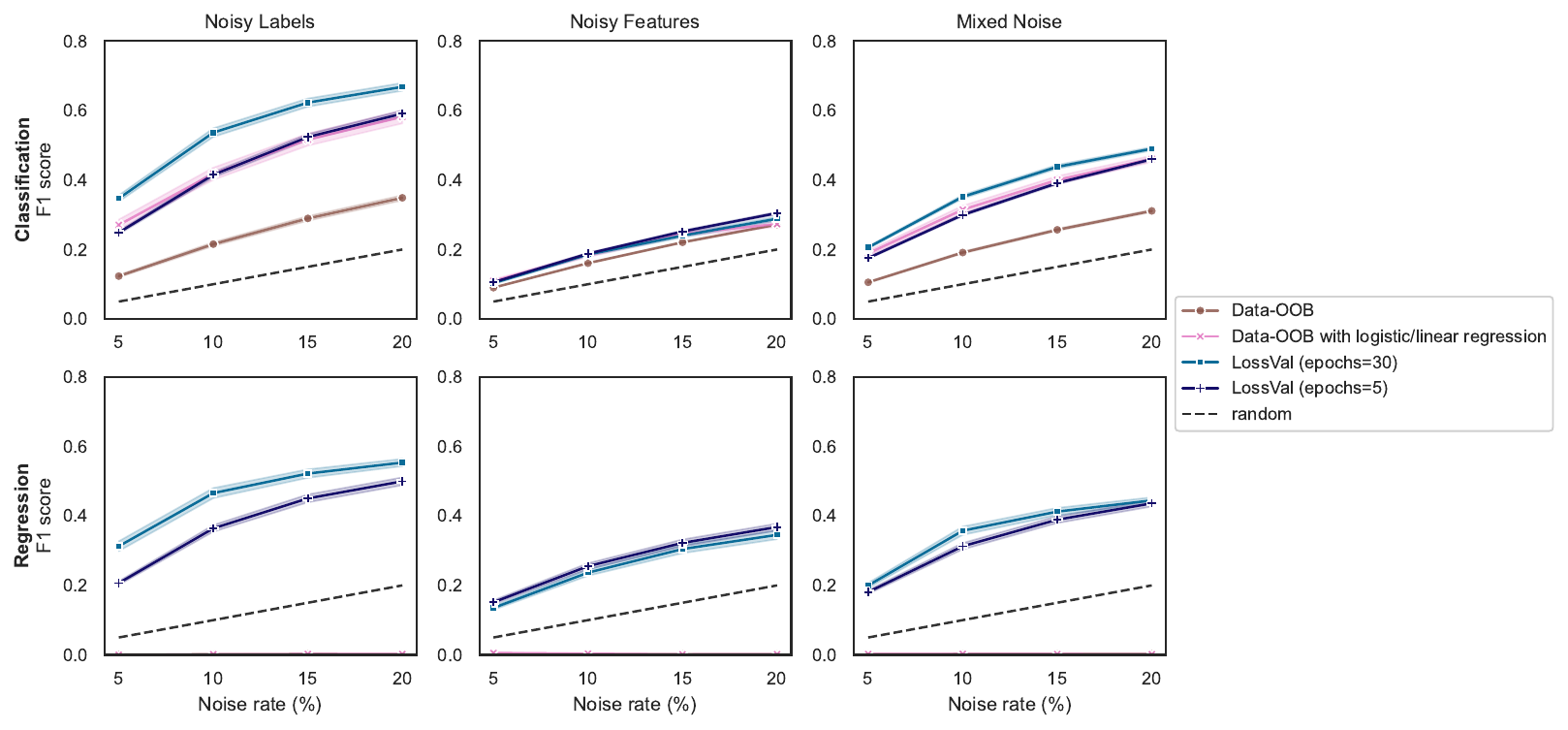}
  \caption{Noisy sample detection comparing Data-OOB with MLP and Data-OOB with logistic or linear regression. Higher is better.}\label{fig:app:noisy_sample_f1_oob}
\end{figure*}

After finishing our experiments, it seemed like Data-OOB~\citep{kwon_data-oob_2023} performed worse in our experiments than in the results provided by \citet{jiang_opendataval_2023}. Upon investigation, we found that using an MLP instead of logistic regression as the base model for classification tasks leads to a decreased performance in the data valuation. We repeated the experiments for Data-OOB using logistic regression and linear regression as base models for the noisy sample detection. \Cref{fig:app:noisy_sample_oob} and \Cref{fig:app:noisy_sample_f1_oob} show that using logistic regression works much better for Data-OOB than using an MLP or linear regression. Still, LossVal achieves similar or better results compared to Data-OOB for both regression and classification.

\begin{table*}
\centering
\fontsize{8}{9}\selectfont
\begin{minipage}{\linewidth}
\begin{tabularx}{\linewidth}{p{.5em}l@{\hskip 0.75em}|YYY|Y}
\toprule
& & \makecell{Mixed\\Noise (\%)} & \makecell{Noisy\\Features (\%)} &  \makecell{Noisy\\Labels (\%)}  &  \makecell{Overall\\Average (\%)} \\
\midrule

\multirow{10}{*}{\rotatebox{90}{\textbf{Classification}}}
& AME                 &              49.98$\pm$0.27 &              49.81$\pm$0.27 &              50.00$\pm$0.27 &              49.93$\pm$0.16 \\
& Beta Shapley         &              53.32$\pm$0.27 &              50.13$\pm$0.26 &              51.76$\pm$0.27 &              51.74$\pm$0.16 \\
& DVRL                &              53.87$\pm$0.28 &              49.97$\pm$0.27 &              52.07$\pm$0.27 &              51.97$\pm$0.16 \\
& Data Banzhaf        &              50.35$\pm$0.27 &              49.97$\pm$0.27 &              50.04$\pm$0.27 &              50.12$\pm$0.16 \\
& Data-OOB            &              56.52$\pm$0.28 &              50.25$\pm$0.27 &              53.19$\pm$0.27 &              53.32$\pm$0.16 \\
& Data Shapley         &              53.25$\pm$0.27 &              50.12$\pm$0.26 &              51.69$\pm$0.27 &              51.69$\pm$0.16 \\
& Influence Subsample &              50.34$\pm$0.27 &              49.96$\pm$0.27 &              50.09$\pm$0.27 &              50.13$\pm$0.16 \\
& KNN-Shapley         &     \textit{72.49}$\pm$0.26 &  \underline{62.78}$\pm$0.35 &  \underline{68.22}$\pm$0.29 &     \textit{67.83}$\pm$0.18 \\
& LAVA                &              54.75$\pm$0.29 &     \textbf{81.18}$\pm$0.24 &     \textbf{68.76}$\pm$0.24 &  \underline{68.23}$\pm$0.16 \\
& Leave-One-Out       &              50.12$\pm$0.27 &              49.85$\pm$0.27 &              49.96$\pm$0.27 &              49.98$\pm$0.16 \\
& LossVal (epochs=5)  &  \underline{81.61}$\pm$0.25 &              53.96$\pm$0.27 &              67.29$\pm$0.23 &              67.62$\pm$0.16 \\
& LossVal (epochs=30) &     \textbf{83.10}$\pm$0.25 &     \textit{55.13}$\pm$0.28 &     \textit{67.91}$\pm$0.23 &     \textbf{68.72}$\pm$0.16 \\

\midrule 

\multirow{10}{*}{\rotatebox{90}{\textbf{Regression}}}
& AME                 &              49.85$\pm$0.27 &              50.04$\pm$0.27 &              49.87$\pm$0.27 &              49.92$\pm$0.16 \\
& Beta Shapley         &              50.14$\pm$0.27 &              50.10$\pm$0.27 &              50.15$\pm$0.27 &              50.13$\pm$0.16 \\
& DVRL                &              49.66$\pm$0.29 &              50.00$\pm$0.27 &              49.25$\pm$0.27 &              49.63$\pm$0.16 \\
& Data Banzhaf        &              49.85$\pm$0.27 &              50.04$\pm$0.27 &              49.87$\pm$0.27 &              49.92$\pm$0.16 \\
& Data-OOB            &              49.85$\pm$0.27 &              50.04$\pm$0.27 &              49.87$\pm$0.27 &              49.92$\pm$0.16 \\
& Data Shapley         &              50.14$\pm$0.27 &              50.10$\pm$0.27 &              50.15$\pm$0.27 &              50.13$\pm$0.16 \\
& Influence Subsample &              49.85$\pm$0.27 &              50.04$\pm$0.27 &              49.87$\pm$0.27 &              49.92$\pm$0.16 \\
& KNN-Shapley         &     \textit{58.88}$\pm$0.28 &              47.33$\pm$0.33 &              53.11$\pm$0.29 &              53.10$\pm$0.17 \\
& LAVA                &              50.03$\pm$0.27 &     \textbf{83.40}$\pm$0.24 &  \underline{66.90}$\pm$0.23 &  \underline{66.77}$\pm$0.16 \\
& Leave-One-Out       &              49.85$\pm$0.27 &              50.04$\pm$0.27 &              49.87$\pm$0.27 &              49.92$\pm$0.16 \\
& LossVal (epochs=5)  &  \underline{70.39}$\pm$0.25 &  \underline{61.47}$\pm$0.27 &     \textit{66.13}$\pm$0.25 &     \textit{66.00}$\pm$0.15 \\
& LossVal (epochs=30) &     \textbf{74.23}$\pm$0.24 &     \textit{61.18}$\pm$0.28 &     \textbf{67.80}$\pm$0.25 &     \textbf{67.74}$\pm$0.15 \\
\bottomrule
\end{tabularx}
\end{minipage}
\caption{Average of the corruption discovery curves of each data valuation method, averaged over all proportion steps, noise rates, and datasets. The number after $\pm$ indicates the standard error. Higher is better. 
}
\label{tab:app:noisy_sample}
\vspace{-1.5em}
\end{table*}

\begin{table*}[!th]
\caption{Results of the point removal and addition experiment, averaged over all removal rates from $0-50\%$ (left) and additional rates from $5-50\%$ (right) and all classification datasets (top) and regression datasets (bottom). 
The number after $\pm$ indicates the standard error. Lower is better.
}\label{tab:exp2:point_addition}\vskip 0.05in
\centerline{
\centering\fontsize{8}{9}\selectfont
\begin{tabularx}{\textwidth}{l@{\hskip 0.25em}l|@{}Y@{}Y@{}Y@{}@{}Y@{}|@{}Y@{}Y@{}Y@{}@{}Y@{}}
\toprule
& & \multicolumn{4}{c|}{\textbf{Point Removal Experiment}} & \multicolumn{4}{c}{\textbf{Point Addition Experiment}} \\

& &  \makecell{Noisy\\Labels} & \makecell{Noisy\\Features} & \makecell{Mixed\\Noise} &  \makecell{Overall\\Average} & \makecell{Noisy\\Labels} & \makecell{Noisy\\Features} & \makecell{Mixed\\Noise} &  \makecell{Overall\\Average}\\
\midrule
\multirow{12}{*}{\rotatebox{90}{\textbf{Classification}}}
& AME                 &              74.86$\pm$0.10 &              75.31$\pm$0.10 &              75.08$\pm$0.10 &              75.08$\pm$0.06 &             64.96$\pm$0.16 &              65.88$\pm$0.16 &              65.81$\pm$0.16 &              65.55$\pm$0.09 \\
& Beta Shapley         &              73.06$\pm$0.11 &              74.40$\pm$0.10 &              73.91$\pm$0.11 &              73.79$\pm$0.06 &             63.98$\pm$0.15 &              65.58$\pm$0.15 &              65.18$\pm$0.16 &              64.91$\pm$0.09 \\
& DVRL                &              70.41$\pm$0.12 &  \underline{72.37}$\pm$0.11 &     \textit{71.74}$\pm$0.11 &     \textit{71.50}$\pm$0.07 &    \textit{61.33}$\pm$0.16 &     \textit{63.38}$\pm$0.15 &     \textit{62.79}$\pm$0.16 &     \textit{62.50}$\pm$0.09 \\
& Data Banzhaf        &              74.66$\pm$0.10 &              75.25$\pm$0.10 &              74.95$\pm$0.10 &              74.96$\pm$0.06 &             64.71$\pm$0.16 &              65.81$\pm$0.16 &              65.67$\pm$0.16 &              65.40$\pm$0.09 \\
& Data-OOB            &              70.45$\pm$0.12 &     \textit{72.55}$\pm$0.11 &              72.02$\pm$0.11 &              71.67$\pm$0.07 &             62.43$\pm$0.15 &              63.99$\pm$0.15 &              63.66$\pm$0.15 &              63.36$\pm$0.09 \\
& Data Shapley         &              73.32$\pm$0.11 &              74.48$\pm$0.10 &              73.99$\pm$0.10 &              73.93$\pm$0.06 &             64.22$\pm$0.15 &              65.66$\pm$0.15 &              65.27$\pm$0.16 &              65.05$\pm$0.09 \\
& Influence Subsample &              74.74$\pm$0.10 &              75.29$\pm$0.10 &              75.01$\pm$0.10 &              75.02$\pm$0.06 &             64.88$\pm$0.16 &              65.90$\pm$0.16 &              65.68$\pm$0.16 &              65.48$\pm$0.09 \\
& KNN-Shapley         &     \textbf{67.21}$\pm$0.14 &     \textbf{71.38}$\pm$0.13 &     \textbf{69.58}$\pm$0.13 &     \textbf{69.39}$\pm$0.08 &    \textbf{60.95}$\pm$0.14 &  \underline{61.87}$\pm$0.14 &     \textbf{61.67}$\pm$0.14 &     \textbf{61.50}$\pm$0.08 \\
& LAVA                &              71.62$\pm$0.11 &              73.78$\pm$0.11 &              73.28$\pm$0.11 &              72.89$\pm$0.06 & \underline{61.04}$\pm$0.14 &     \textbf{61.85}$\pm$0.15 &  \underline{61.97}$\pm$0.15 &  \underline{61.62}$\pm$0.08 \\
& Leave-One-Out       &              74.89$\pm$0.10 &              75.33$\pm$0.10 &              75.06$\pm$0.10 &              75.10$\pm$0.06 &             64.99$\pm$0.16 &              65.88$\pm$0.16 &              65.83$\pm$0.16 &              65.57$\pm$0.09 \\
& LossVal (epochs=5)  &  \underline{68.64}$\pm$0.17 &              73.15$\pm$0.13 &  \underline{71.22}$\pm$0.15 &  \underline{71.00}$\pm$0.09 &             65.23$\pm$0.15 &              65.87$\pm$0.15 &              65.17$\pm$0.15 &              65.42$\pm$0.08 \\
& LossVal (epochs=30) &     \textit{69.86}$\pm$0.17 &              73.57$\pm$0.12 &              72.01$\pm$0.14 &              71.81$\pm$0.08 &             65.29$\pm$0.15 &              66.21$\pm$0.15 &              65.46$\pm$0.15 &              65.65$\pm$0.09 \\

\midrule 

\multirow{12}{*}{\rotatebox{90}{\textbf{Regression}}}
& AME                 &              -0.143$\pm$0.001 &              -0.128$\pm$0.001 &              -0.129$\pm$0.001 &              -0.133$\pm$0.001 &              -0.011$\pm$0.000 &              -0.009$\pm$0.000 &              -0.010$\pm$0.000 &              -0.010$\pm$0.000 \\
& Beta Shapley         &              -0.146$\pm$0.001 &              -0.122$\pm$0.001 &              -0.134$\pm$0.001 &              -0.134$\pm$0.001 &              -0.011$\pm$0.000 &              -0.009$\pm$0.000 &              -0.010$\pm$0.000 &              -0.010$\pm$0.000 \\
& DVRL                &     \textbf{-0.173}$\pm$0.002 &     \textbf{-0.151}$\pm$0.002 &     \textbf{-0.154}$\pm$0.002 &     \textbf{-0.160}$\pm$0.001 &     \textbf{-0.030}$\pm$0.001 &     \textbf{-0.021}$\pm$0.001 &     \textbf{-0.025}$\pm$0.001 &     \textbf{-0.025}$\pm$0.000 \\
& Data Banzhaf        &              -0.143$\pm$0.001 &              -0.128$\pm$0.001 &              -0.129$\pm$0.001 &              -0.133$\pm$0.001 &              -0.011$\pm$0.000 &              -0.009$\pm$0.000 &              -0.010$\pm$0.000 &              -0.010$\pm$0.000 \\
& Data-OOB            &              -0.143$\pm$0.001 &              -0.128$\pm$0.001 &              -0.129$\pm$0.001 &              -0.133$\pm$0.001 &              -0.011$\pm$0.000 &              -0.009$\pm$0.000 &              -0.010$\pm$0.000 &              -0.010$\pm$0.000 \\
& Data Shapley         &              -0.146$\pm$0.001 &              -0.122$\pm$0.001 &              -0.134$\pm$0.001 &              -0.134$\pm$0.001 &              -0.011$\pm$0.000 &              -0.009$\pm$0.000 &              -0.010$\pm$0.000 &              -0.010$\pm$0.000 \\
& Influence Subsample &              -0.143$\pm$0.001 &              -0.128$\pm$0.001 &              -0.129$\pm$0.001 &              -0.133$\pm$0.001 &              -0.011$\pm$0.000 &              -0.009$\pm$0.000 &              -0.010$\pm$0.000 &              -0.010$\pm$0.000 \\
& KNN-Shapley         &     \textit{-0.163}$\pm$0.002 &     \textit{-0.137}$\pm$0.001 &              -0.140$\pm$0.001 &     \textit{-0.147}$\pm$0.001 &     \textit{-0.012}$\pm$0.000 &     \textit{-0.011}$\pm$0.000 &     \textit{-0.011}$\pm$0.000 &     \textit{-0.012}$\pm$0.000 \\
& LAVA                &              -0.148$\pm$0.002 &  \underline{-0.148}$\pm$0.001 &     \textit{-0.142}$\pm$0.001 &              -0.146$\pm$0.001 &  \underline{-0.016}$\pm$0.000 &  \underline{-0.013}$\pm$0.000 &  \underline{-0.015}$\pm$0.000 &  \underline{-0.015}$\pm$0.000 \\
& Leave-One-Out       &              -0.143$\pm$0.001 &              -0.128$\pm$0.001 &              -0.129$\pm$0.001 &              -0.133$\pm$0.001 &              -0.011$\pm$0.000 &              -0.009$\pm$0.000 &              -0.010$\pm$0.000 &              -0.010$\pm$0.000 \\
& LossVal (epochs=5)  &              -0.152$\pm$0.002 &              -0.129$\pm$0.001 &              -0.135$\pm$0.001 &              -0.139$\pm$0.001 &              -0.006$\pm$0.000 &              -0.006$\pm$0.000 &              -0.006$\pm$0.000 &              -0.006$\pm$0.000 \\
& LossVal (epochs=30) &  \underline{-0.165}$\pm$0.002 &              -0.135$\pm$0.001 &  \underline{-0.144}$\pm$0.001 &  \underline{-0.148}$\pm$0.001 &              -0.007$\pm$0.000 &              -0.007$\pm$0.000 &              -0.007$\pm$0.000 &              -0.007$\pm$0.000 \\

\bottomrule
\end{tabularx}}
\vskip -0.1in
\end{table*}

\begin{table*}
\caption{Average of the noisy sample detection F1 scores on the classification task, averaged over all noise rates. The number after $\pm$ indicates the standard error. Higher is better. 
}\label{tab:app:f1_scores_classification}
\centering
\fontsize{8}{9}\selectfont
\begin{tabularx}{1\textwidth}{p{.5em}l@{\hskip 0.75em}|YYY|Y}
\toprule
& &  \makecell{Noisy\\Labels} & \makecell{Noisy\\Features} & \makecell{Mixed\\Noise} &  \makecell{Overall\\Average} \\

\midrule
\multirow{10}{*}{\rotatebox{90}{\textbf{2dplanes}}}
& AME                 &              0.069$\pm$.012 &              0.091$\pm$.012 &              0.080$\pm$.011 &              0.080$\pm$.007 \\
& DVRL                &              0.192$\pm$.008 &              0.191$\pm$.008 &              0.187$\pm$.008 &              0.190$\pm$.004 \\
& Data Banzhaf        &              0.196$\pm$.007 &              0.194$\pm$.007 &              0.191$\pm$.008 &              0.193$\pm$.004 \\
& Data-OOB            &              0.191$\pm$.007 &              0.193$\pm$.008 &              0.193$\pm$.007 &              0.192$\pm$.004 \\
& Influence Subsample &              0.193$\pm$.007 &              0.189$\pm$.008 &              0.189$\pm$.007 &              0.190$\pm$.004 \\
& KNN-Shapley         &     \textit{0.344}$\pm$.013 &  \underline{0.274}$\pm$.010 &     \textit{0.316}$\pm$.011 &              0.311$\pm$.007 \\
& LAVA                &              0.213$\pm$.007 &     \textbf{0.434}$\pm$.016 &              0.297$\pm$.011 &     \textit{0.315}$\pm$.009 \\
& Leave-One-Out       &              0.183$\pm$.008 &              0.163$\pm$.010 &              0.179$\pm$.009 &              0.175$\pm$.005 \\
& LossVal (epochs=5)  &  \underline{0.462}$\pm$.014 &              0.236$\pm$.008 &  \underline{0.366}$\pm$.012 &  \underline{0.356}$\pm$.009 \\
& LossVal (epochs=30) &     \textbf{0.592}$\pm$.014 &     \textit{0.243}$\pm$.008 &     \textbf{0.427}$\pm$.012 &     \textbf{0.423}$\pm$.011 \\

\midrule
\multirow{10}{*}{\rotatebox{90}{\textbf{electricity}}}
& AME                 &              0.067$\pm$.011 &              0.090$\pm$.013 &              0.079$\pm$.012 &              0.079$\pm$.007 \\
& DVRL                &              0.188$\pm$.008 &              0.186$\pm$.008 &              0.192$\pm$.008 &              0.189$\pm$.005 \\
& Data Banzhaf        &              0.194$\pm$.007 &              0.193$\pm$.008 &              0.193$\pm$.008 &              0.193$\pm$.004 \\
& Data-OOB            &              0.196$\pm$.007 &              0.196$\pm$.007 &              0.194$\pm$.007 &              0.195$\pm$.004 \\
& Influence Subsample &              0.193$\pm$.008 &              0.192$\pm$.008 &              0.191$\pm$.008 &              0.192$\pm$.004 \\
& KNN-Shapley         &     \textit{0.289}$\pm$.010 &     \textbf{0.216}$\pm$.008 &     \textit{0.255}$\pm$.009 &     \textit{0.253}$\pm$.006 \\
& LAVA                &              0.023$\pm$.007 &              0.060$\pm$.014 &              0.034$\pm$.008 &              0.039$\pm$.006 \\
& Leave-One-Out       &              0.180$\pm$.010 &              0.165$\pm$.009 &              0.176$\pm$.008 &              0.174$\pm$.005 \\
& LossVal (epochs=5)  &  \underline{0.331}$\pm$.011 &  \underline{0.207}$\pm$.008 &  \underline{0.271}$\pm$.010 &  \underline{0.270}$\pm$.006 \\
& LossVal (epochs=30) &     \textbf{0.359}$\pm$.012 &     \textit{0.201}$\pm$.008 &     \textbf{0.282}$\pm$.010 &     \textbf{0.281}$\pm$.007 \\

\midrule
\multirow{10}{*}{\rotatebox{90}{\textbf{fried}}}
& AME                 &              0.077$\pm$.011 &              0.084$\pm$.012 &              0.091$\pm$.012 &              0.084$\pm$.007 \\
& DVRL                &              0.190$\pm$.008 &              0.191$\pm$.008 &              0.190$\pm$.008 &              0.190$\pm$.004 \\
& Data Banzhaf        &              0.189$\pm$.008 &              0.195$\pm$.008 &              0.190$\pm$.008 &              0.191$\pm$.004 \\
& Data-OOB            &              0.190$\pm$.007 &              0.192$\pm$.007 &              0.196$\pm$.007 &              0.193$\pm$.004 \\
& Influence Subsample &              0.195$\pm$.007 &              0.194$\pm$.007 &              0.193$\pm$.007 &              0.194$\pm$.004 \\
& KNN-Shapley         &     \textit{0.322}$\pm$.012 &  \underline{0.264}$\pm$.010 &     \textit{0.295}$\pm$.010 &              0.294$\pm$.006 \\
& LAVA                &              0.200$\pm$.007 &     \textbf{0.419}$\pm$.016 &              0.284$\pm$.010 &     \textit{0.301}$\pm$.008 \\
& Leave-One-Out       &              0.181$\pm$.009 &              0.181$\pm$.008 &              0.178$\pm$.009 &              0.180$\pm$.005 \\
& LossVal (epochs=5)  &  \underline{0.437}$\pm$.013 &              0.213$\pm$.008 &  \underline{0.331}$\pm$.011 &  \underline{0.327}$\pm$.008 \\
& LossVal (epochs=30) &     \textbf{0.535}$\pm$.014 &     \textit{0.224}$\pm$.008 &     \textbf{0.384}$\pm$.012 &     \textbf{0.381}$\pm$.010 \\
\midrule

\multirow{10}{*}{\rotatebox{90}{\textbf{MiniBooNE}}}
& AME                 &              0.093$\pm$.012 &              0.083$\pm$.011 &              0.088$\pm$.012 &              0.088$\pm$.007 \\
& DVRL                &              0.185$\pm$.008 &              0.199$\pm$.011 &              0.190$\pm$.008 &              0.191$\pm$.005 \\
& Data Banzhaf        &              0.194$\pm$.008 &              0.195$\pm$.008 &              0.194$\pm$.007 &              0.194$\pm$.004 \\
& Data-OOB            &              0.209$\pm$.009 &              0.193$\pm$.008 &              0.194$\pm$.007 &              0.199$\pm$.005 \\
& Influence Subsample &              0.193$\pm$.008 &              0.192$\pm$.007 &              0.193$\pm$.007 &              0.193$\pm$.004 \\
& KNN-Shapley         &  \underline{0.374}$\pm$.013 &     \textbf{0.368}$\pm$.012 &     \textbf{0.365}$\pm$.014 &     \textbf{0.369}$\pm$.008 \\
& LAVA                &              0.046$\pm$.010 &              0.094$\pm$.023 &              0.100$\pm$.018 &              0.080$\pm$.010 \\
& Leave-One-Out       &              0.139$\pm$.010 &              0.171$\pm$.009 &              0.170$\pm$.009 &              0.160$\pm$.006 \\
& LossVal (epochs=5)  &     \textit{0.355}$\pm$.014 &  \underline{0.228}$\pm$.009 &     \textit{0.292}$\pm$.010 &     \textit{0.292}$\pm$.007 \\
& LossVal (epochs=30) &     \textbf{0.421}$\pm$.016 &     \textit{0.203}$\pm$.011 &  \underline{0.321}$\pm$.011 &  \underline{0.315}$\pm$.009 \\

\midrule
\multirow{10}{*}{\rotatebox{90}{\textbf{nomao}}}
& AME                 &              0.063$\pm$.011 &              0.059$\pm$.010 &              0.065$\pm$.011 &              0.062$\pm$.006 \\
& DVRL                &              0.309$\pm$.015 &              0.186$\pm$.007 &              0.246$\pm$.011 &              0.247$\pm$.007 \\
& Data Banzhaf        &              0.195$\pm$.008 &              0.165$\pm$.009 &              0.183$\pm$.009 &              0.181$\pm$.005 \\
& Data-OOB            &              0.364$\pm$.011 &              0.167$\pm$.006 &              0.272$\pm$.009 &              0.268$\pm$.007 \\
& Influence Subsample &              0.195$\pm$.008 &              0.178$\pm$.009 &              0.189$\pm$.009 &              0.187$\pm$.005 \\
& KNN-Shapley         &     \textit{0.483}$\pm$.013 &  \underline{0.200}$\pm$.009 &     \textit{0.302}$\pm$.009 &     \textit{0.328}$\pm$.009 \\
& LAVA                &              0.051$\pm$.004 &     \textbf{0.280}$\pm$.029 &              0.205$\pm$.020 &              0.179$\pm$.013 \\
& Leave-One-Out       &              0.210$\pm$.010 &     \textit{0.193}$\pm$.010 &              0.202$\pm$.010 &              0.202$\pm$.006 \\
& LossVal (epochs=5)  &  \underline{0.522}$\pm$.014 &              0.163$\pm$.006 &  \underline{0.353}$\pm$.011 &  \underline{0.346}$\pm$.011 \\
& LossVal (epochs=30) &     \textbf{0.637}$\pm$.014 &              0.124$\pm$.005 &     \textbf{0.395}$\pm$.011 &     \textbf{0.385}$\pm$.014 \\

\midrule
\multirow{10}{*}{\rotatebox{90}{\textbf{pol}}}
& AME                 &              0.075$\pm$.013 &              0.006$\pm$.002 &              0.012$\pm$.005 &              0.031$\pm$.005 \\
& DVRL                &              0.295$\pm$.011 &              0.170$\pm$.007 &              0.239$\pm$.009 &              0.235$\pm$.006 \\
& Data Banzhaf        &              0.133$\pm$.013 &              0.029$\pm$.007 &              0.075$\pm$.012 &              0.079$\pm$.007 \\
& Data-OOB            &              0.316$\pm$.011 &              0.174$\pm$.007 &              0.248$\pm$.009 &              0.246$\pm$.006 \\
& Influence Subsample &              0.134$\pm$.013 &              0.019$\pm$.006 &              0.065$\pm$.012 &              0.073$\pm$.007 \\
& KNN-Shapley         &     \textit{0.319}$\pm$.011 &              0.176$\pm$.006 &              0.258$\pm$.010 &              0.251$\pm$.006 \\
& LAVA                &              0.064$\pm$.004 &     \textbf{0.686}$\pm$.016 &  \underline{0.400}$\pm$.015 &     \textit{0.383}$\pm$.017 \\
& Leave-One-Out       &              0.143$\pm$.014 &              0.027$\pm$.008 &              0.088$\pm$.013 &              0.086$\pm$.007 \\
& LossVal (epochs=5)  &  \underline{0.555}$\pm$.016 &     \textit{0.228}$\pm$.009 &     \textit{0.375}$\pm$.012 &  \underline{0.386}$\pm$.011 \\
& LossVal (epochs=30) &     \textbf{0.712}$\pm$.014 &  \underline{0.231}$\pm$.009 &     \textbf{0.420}$\pm$.012 &     \textbf{0.454}$\pm$.013 \\

\bottomrule
\end{tabularx}
\end{table*}

\begin{table*}
\caption{Average of the noisy sample detection F1 scores on the regression task, averaged over all noise rates. The number after $\pm$ indicates the standard error. Higher is better.}
\label{tab:app:f1_scores_regression}
\centering
\fontsize{8}{9}\selectfont
\begin{tabularx}{1\textwidth}{p{.5em}l@{\hskip 0.75em}|YYY|Y}
\toprule
& &  \makecell{Noisy\\Labels} & \makecell{Noisy\\Features} & \makecell{Mixed\\Noise} &  \makecell{Overall\\Average} \\

\midrule
\multirow{10}{*}{\rotatebox{90}{\textbf{cpu\_activity}}}
& AME                 &              0.002$\pm$.001 &              0.002$\pm$.001 &              0.002$\pm$.001 &              0.002$\pm$.000 \\
& DVRL                &              0.240$\pm$.015 &              0.199$\pm$.011 &              0.227$\pm$.013 &              0.222$\pm$.008 \\
& Data Banzhaf        &              0.002$\pm$.001 &              0.002$\pm$.001 &              0.002$\pm$.001 &              0.002$\pm$.000 \\
& Data-OOB            &              0.002$\pm$.001 &              0.002$\pm$.001 &              0.002$\pm$.001 &              0.002$\pm$.000 \\
& Influence Subsample &              0.002$\pm$.001 &              0.002$\pm$.001 &              0.002$\pm$.001 &              0.002$\pm$.000 \\
& KNN-Shapley         &     \textit{0.268}$\pm$.010 &     \textit{0.269}$\pm$.010 &     \textit{0.269}$\pm$.011 &     \textit{0.269}$\pm$.006 \\
& LAVA                &              0.012$\pm$.002 &              0.029$\pm$.008 &              0.021$\pm$.003 &              0.021$\pm$.003 \\
& Leave-One-Out       &              0.002$\pm$.001 &              0.002$\pm$.001 &              0.002$\pm$.001 &              0.002$\pm$.000 \\
& LossVal (epochs=5)  &  \underline{0.470}$\pm$.014 &     \textbf{0.409}$\pm$.009 &  \underline{0.464}$\pm$.011 &  \underline{0.448}$\pm$.007 \\
& LossVal (epochs=30) &     \textbf{0.638}$\pm$.013 &  \underline{0.367}$\pm$.009 &     \textbf{0.474}$\pm$.010 &     \textbf{0.493}$\pm$.009 \\

\midrule
\multirow{10}{*}{\rotatebox{90}{\textbf{kin8nm}}}
& AME                 &              0.001$\pm$.000 &              0.002$\pm$.001 &              0.002$\pm$.001 &              0.002$\pm$.000 \\
& DVRL                &     \textit{0.217}$\pm$.013 &              0.208$\pm$.010 &              0.223$\pm$.011 &              0.216$\pm$.006 \\
& Data Banzhaf        &              0.001$\pm$.000 &              0.002$\pm$.001 &              0.002$\pm$.001 &              0.002$\pm$.000 \\
& Data-OOB            &              0.001$\pm$.000 &              0.002$\pm$.001 &              0.002$\pm$.001 &              0.002$\pm$.000 \\
& Influence Subsample &              0.001$\pm$.000 &              0.002$\pm$.001 &              0.002$\pm$.001 &              0.002$\pm$.000 \\
& KNN-Shapley         &              0.001$\pm$.000 &              0.002$\pm$.001 &              0.003$\pm$.001 &              0.002$\pm$.000 \\
& LAVA                &              0.186$\pm$.007 &     \textbf{0.403}$\pm$.014 &  \underline{0.271}$\pm$.010 &  \underline{0.287}$\pm$.008 \\
& Leave-One-Out       &              0.001$\pm$.000 &              0.002$\pm$.001 &              0.002$\pm$.001 &              0.002$\pm$.000 \\
& LossVal (epochs=5)  &  \underline{0.292}$\pm$.009 &     \textit{0.247}$\pm$.008 &     \textit{0.270}$\pm$.009 &     \textit{0.270}$\pm$.005 \\
& LossVal (epochs=30) &     \textbf{0.409}$\pm$.012 &  \underline{0.315}$\pm$.010 &     \textbf{0.364}$\pm$.011 &     \textbf{0.363}$\pm$.007 \\

\midrule
\multirow{10}{*}{\rotatebox{90}{\textbf{pumadyn32nh}}}
& AME                 &              0.002$\pm$.001 &              0.002$\pm$.001 &              0.002$\pm$.001 &              0.002$\pm$.000 \\
& DVRL                &     \textit{0.201}$\pm$.009 &     \textit{0.196}$\pm$.008 &              0.196$\pm$.008 &              0.198$\pm$.005 \\
& Data Banzhaf        &              0.002$\pm$.001 &              0.002$\pm$.001 &              0.002$\pm$.001 &              0.002$\pm$.000 \\
& Data-OOB            &              0.002$\pm$.001 &              0.002$\pm$.001 &              0.002$\pm$.001 &              0.002$\pm$.000 \\
& Influence Subsample &              0.002$\pm$.001 &              0.002$\pm$.001 &              0.002$\pm$.001 &              0.002$\pm$.000 \\
& KNN-Shapley         &              0.002$\pm$.001 &              0.004$\pm$.003 &              0.006$\pm$.003 &              0.004$\pm$.001 \\
& LAVA                &              0.193$\pm$.007 &     \textbf{0.735}$\pm$.020 &     \textbf{0.403}$\pm$.017 &     \textbf{0.444}$\pm$.016 \\
& Leave-One-Out       &              0.002$\pm$.001 &              0.002$\pm$.001 &              0.002$\pm$.001 &              0.002$\pm$.000 \\
& LossVal (epochs=5)  &  \underline{0.203}$\pm$.007 &              0.192$\pm$.007 &     \textit{0.199}$\pm$.007 &     \textit{0.198}$\pm$.004 \\
& LossVal (epochs=30) &     \textbf{0.248}$\pm$.007 &  \underline{0.203}$\pm$.007 &  \underline{0.225}$\pm$.008 &  \underline{0.225}$\pm$.004 \\

\midrule
\multirow{10}{*}{\rotatebox{90}{\textbf{superconductivity}}}
& AME                 &              0.002$\pm$.000 &              0.001$\pm$.000 &              0.002$\pm$.001 &              0.001$\pm$.000 \\
& DVRL                &     \textit{0.256}$\pm$.014 &              0.171$\pm$.010 &              0.207$\pm$.008 &              0.211$\pm$.007 \\
& Data Banzhaf        &              0.002$\pm$.000 &              0.001$\pm$.000 &              0.002$\pm$.001 &              0.001$\pm$.000 \\
& Data-OOB            &              0.002$\pm$.000 &              0.001$\pm$.000 &              0.002$\pm$.001 &              0.001$\pm$.000 \\
& Influence Subsample &              0.002$\pm$.000 &              0.001$\pm$.000 &              0.002$\pm$.001 &              0.001$\pm$.000 \\
& KNN-Shapley         &              0.222$\pm$.009 &  \underline{0.227}$\pm$.009 &     \textit{0.225}$\pm$.009 &     \textit{0.225}$\pm$.005 \\
& LAVA                &              0.126$\pm$.006 &     \textbf{0.652}$\pm$.014 &     \textbf{0.437}$\pm$.010 &     \textbf{0.405}$\pm$.014 \\
& Leave-One-Out       &              0.002$\pm$.000 &              0.001$\pm$.000 &              0.002$\pm$.001 &              0.001$\pm$.000 \\
& LossVal (epochs=5)  &  \underline{0.331}$\pm$.010 &     \textit{0.177}$\pm$.006 &  \underline{0.251}$\pm$.008 &  \underline{0.253}$\pm$.006 \\
& LossVal (epochs=30) &     \textbf{0.360}$\pm$.011 &              0.063$\pm$.003 &              0.220$\pm$.007 &              0.215$\pm$.008 \\

\midrule
\multirow{10}{*}{\rotatebox{90}{\textbf{wave\_energy}}}
& AME                 &              0.002$\pm$.001 &              0.002$\pm$.001 &              0.002$\pm$.001 &              0.002$\pm$.000 \\
& DVRL                &     \textit{0.263}$\pm$.017 &              0.228$\pm$.012 &              0.224$\pm$.014 &              0.238$\pm$.008 \\
& Data Banzhaf        &              0.002$\pm$.001 &              0.002$\pm$.001 &              0.002$\pm$.001 &              0.002$\pm$.000 \\
& Data-OOB            &              0.002$\pm$.001 &              0.002$\pm$.001 &              0.002$\pm$.001 &              0.002$\pm$.000 \\
& Influence Subsample &              0.002$\pm$.001 &              0.002$\pm$.001 &              0.002$\pm$.001 &              0.002$\pm$.000 \\
& KNN-Shapley         &              0.002$\pm$.001 &              0.002$\pm$.001 &              0.002$\pm$.001 &              0.002$\pm$.000 \\
& LAVA                &              0.206$\pm$.008 &     \textbf{0.436}$\pm$.027 &     \textit{0.256}$\pm$.010 &     \textit{0.299}$\pm$.011 \\
& Leave-One-Out       &              0.002$\pm$.001 &              0.002$\pm$.001 &              0.002$\pm$.001 &              0.002$\pm$.000 \\
& LossVal (epochs=5)  &  \underline{0.499}$\pm$.015 &  \underline{0.419}$\pm$.014 &  \underline{0.454}$\pm$.014 &  \underline{0.458}$\pm$.008 \\
& LossVal (epochs=30) &     \textbf{0.660}$\pm$.008 &     \textit{0.401}$\pm$.015 &     \textbf{0.517}$\pm$.015 &     \textbf{0.526}$\pm$.010 \\

\midrule
\multirow{10}{*}{\rotatebox{90}{\textbf{white\_wine}}}
& AME                 &              0.002$\pm$.001 &              0.002$\pm$.001 &              0.002$\pm$.000 &              0.002$\pm$.000 \\
& DVRL                &              0.306$\pm$.024 &     \textit{0.184}$\pm$.008 &              0.230$\pm$.012 &              0.240$\pm$.010 \\
& Data Banzhaf        &              0.002$\pm$.001 &              0.002$\pm$.001 &              0.002$\pm$.000 &              0.002$\pm$.000 \\
& Data-OOB            &              0.002$\pm$.001 &              0.002$\pm$.001 &              0.002$\pm$.000 &              0.002$\pm$.000 \\
& Influence Subsample &              0.002$\pm$.001 &              0.002$\pm$.001 &              0.002$\pm$.000 &              0.002$\pm$.000 \\
& KNN-Shapley         &     \textit{0.428}$\pm$.015 &              0.162$\pm$.008 &     \textit{0.288}$\pm$.010 &     \textit{0.292}$\pm$.009 \\
& LAVA                &              0.037$\pm$.005 &     \textbf{0.233}$\pm$.023 &              0.139$\pm$.015 &              0.136$\pm$.010 \\
& Leave-One-Out       &              0.002$\pm$.001 &              0.002$\pm$.001 &              0.002$\pm$.000 &              0.002$\pm$.000 \\
& LossVal (epochs=5)  &     \textbf{0.486}$\pm$.015 &  \underline{0.202}$\pm$.007 &     \textbf{0.340}$\pm$.011 &     \textbf{0.343}$\pm$.009 \\
& LossVal (epochs=30) &  \underline{0.466}$\pm$.015 &              0.184$\pm$.007 &  \underline{0.321}$\pm$.011 &  \underline{0.324}$\pm$.009 \\

\bottomrule
\end{tabularx}
\end{table*}

\section{Running Times}\label{app:sec:runtime}
\begin{table}[!th]
\caption{Average runtime on the \textit{2dplanes} dataset (classification). Some overhead from the benchmark code around it should be expected, but should be comparable across all methods.}\label{tab:app:runtime}\vskip 0.05in
\centering\fontsize{8}{9}\selectfont
\begin{tabular}{l@{\hskip 2em}|@{\hskip 2em}l}
\toprule
                    & \makecell{1000 Datapoints} \\
\midrule
AME                  &  8min 28s 983ms \\
Beta Shapley         &  3min 30s 701ms \\
DVRL                 &  0min 33s 25ms \\
Data Banzhaf         &  2min 07s 145ms \\
Data-OOB             &  4min 05s 226ms \\
Data Shapley         &  3min 16s 276ms \\
Influence Subsample  &  2min 51s 927ms \\
KNN-Shapley          &  0min 00s 160ms \\
LAVA                 &  0min 00s 104ms \\
Leave-One-Out        &  4min 04s 954ms \\
LossVal (epochs=5)   &  0min 01s 819ms \\
LossVal (epochs=30)  &  0min 10s 617ms \\
\bottomrule
\end{tabular}
\vskip -0.1in
\end{table}

\Cref{tab:app:runtime} shows a runtime comparison between the baselines used in this study and LossVal. 
The measurement was repeated five times on a RTX 3060 GPU. 
Note that LAVA and KNN-Shapley were executed on an 8-core CPU instead of the GPU, because they are model-free and do not train an MLP as the other methods do. 
Accordingly, observe that they are faster than the other methods. Aside from those two, LossVal with 5 and LossVal with 30 epochs are significantly faster than the baselines. 

We observe that Data Shapley is faster than Data-OOB, because the implementation of Data Shapley uses an approximation instead of calculating the exact Shapley values. The table also shows that the real-world performance of Data-OOB, Leave-One-Out, and Influence Subsample differ, even though have the same number of training runs. This stems from the fact, that they use subsets of different size for retraining, affecting the duration of each training run.

\section{Extended Related Work}
\label{sec:extended-related-work}

We compare our methods to representative methods from the main branches of Data Valuation methods, as described in \Cref{sec:related_work}. We use them as strong baselines to show the general feasibility of our approach. However, there is a multitude of data valuation approaches that we left out in the comparison for feasibility, that are still worth mentioning here. 

There is a range of approaches to extend Data Shapley, either to make it more efficient or to achieve better results on the benchmarks~\citep{schoch_cs-shapley_2022, panda_fw-shapley_2024, cai_chg_2024, schoch_cs-shapley_2022, zheng_towards_2024, pombal_fairness-aware_2023}. 

None of them performs so much better than Data Shapley and Beta Shapley that we felt the need to include them in the benchmark. Other approaches try to employ other useful ideas from game theory, like the Banzhaf value (which is included in the comparison) and the Winter value~\citep{chi_precedence-constrained_2024}.

Furthermore, there are some model-free approaches similar to LAVA\citep{just_lava_2023, lin_distributionally_2024, kessler_sava_2024} and approaches that apply data valuation without the use of a validation set~\citep{jahagirdar_data_2024}. Other approaches apply methods similar to data valuation to machine learning models, datasets, data clusters, or distributions~\citep{sun_2d-oob_2024, tarun_ecoval_2024, xu_data_2024, xu_model_2023, yona_whos_2021}. Instead of assigning a value to each data point, they assign a value to each model, dataset, data cluster, or distribution, respectively. Some modifications to data valuation allow a joint valuation of model and data point, or data points and data \enquote{cells}.

Influence-based approaches, like Influence Functions, are generally quite inefficient. We used Influence Subsample to approximate the exact calculations of Influence Functions. Other influence-based approaches, such as Gradient Sketching~\citep{schioppa_gradient_2024} were left out.

Some notable methods left out in the comparison are Simfluence~\citep{guu_simfluence_2023} and Neural Dynamic Data Valuation (NDDV)~\citep{liang_neural_2024}. Simfluence executes multiple training runs and trains a second model on loss over time of a training run, based on the order in which the data points are sampled. 
The second model is used to simulate more training runs, which can then be used to estimate how important each data point is. NDDV use optimal control strategies to understand the importance of data points without needing to retrain the utility function.

Aside from that, some interesting applications of data valuation are described in the literature. \citet{nerini_value_2024} study the applications of data valuation for graph-based data and data markets. An increasing number of papers focuses on the use of data valuation in an economic context or in data markets~\citep{tian_private_2023, agarwal_marketplace_2019, chen_model-based_2017, chen_towards_2019, li_theory_2015, raskar_data_2019, mieth_data_2024}. \citet{wang_adversarial_2024} demonstrate Data Valuation methods can be attacked in an adversarial manner. \citet{tian_derdava_2024} study how data valuation can be made more robust to deletion of data points.

\end{document}